\documentclass[manuscript, screen]{jair}

\usepackage{amssymb}  % for \lnot, \land
\usepackage{amsmath}
\usepackage{graphicx}
\usepackage{algpseudocode}
\usepackage{xcolor}  % colors
\usepackage[table]{xcolor} 
\usepackage{adjustbox}
\usepackage{algorithm}
\usepackage{rotating}
\usepackage{float,lscape}
\usepackage{subcaption}

\setcopyright{cc}
\copyrightyear{2025}
\JAIRTrack{Integration of Logical Constraints in Deep Learning} 
\acmVolume{4}
\acmArticle{6}
\acmMonth{8}
\acmYear{2025}

\RequirePackage[
  datamodel=acmdatamodel,
  style=acmauthoryear,
  backend=biber,
  giveninits=true,
  uniquename=init
  ]{biblatex}

\begin{document}

\pretolerance = 10000
%\title{JAIR Example Template}
\title[Guided by Stars]{Guided by Stars: Interpretable Concept Learning Over Time Series via Temporal Logic Semantics}

%% used to denote shared contribution to the research and/or corresponding author.

\author{Irene Ferfoglia}
\authornote{Corresponding authors.}
\orcid{0000-0003-1585-6576}
\email{irene.ferfo@gmail.com}
\affiliation{%
  \institution{Università degli Studi di Trieste}
  \city{Trieste}
  \country{Italy}
}

\author{Simone Silvetti}
\orcid{0000-0001-8048-9317}
\affiliation{%
  \institution{Università degli Studi di Trieste}
  \city{Trieste}
  \country{Italy}}

\author{Gaia Saveri}
\orcid{0009-0003-2948-7705}
\affiliation{%
  \institution{Università degli Studi di Trieste}
  \city{Trieste}
  \country{Italy}
}

\author{Laura Nenzi}
\orcid{0000-0003-2263-9342}
\affiliation{%
  \institution{Università degli Studi di Trieste}
  \city{Trieste}
  \country{Italy}
}

\author{Luca Bortolussi\small{*}}
\orcid{0000-0001-8874-4001}
\email{lbortolussi@units.it}
\affiliation{%
  \institution{Università degli Studi di Trieste}
  \city{Trieste}
  \country{Italy}
}
%% The short list of authors must be made of the list of all authors' lastnames.
\renewcommand{\shortauthors}{Ferfoglia, Silvetti, Saveri, Nenzi \& Bortolussi}
%% If this is too long and overlaps other information printed in the page headers, use
%\renewcommand{\shortauthors}{Xu et al.}

\begin{abstract}
Time series classification is a task of paramount importance, as this kind of data often arises in safety-critical applications. However, it is typically tackled with black-box deep learning methods, making it hard for humans to understand the rationale behind their output. 
To take on this challenge, we propose a novel approach, STELLE (Signal Temporal logic Embedding for Logically-grounded Learning and Explanation), a neuro-symbolic framework that unifies classification and explanation through direct embedding of trajectories into a space of temporal logic concepts. By introducing a novel STL-inspired kernel that maps raw time series to their alignment with predefined STL formulae, our model jointly optimises accuracy and interpretability, as each prediction is accompanied by the most relevant logical concepts that characterise it.
% whose key resides in embed both time series trajectories and Signal Temporal Logic (STL) formulae into a shared semantic space using a dual kernel-based mechanism. In particular, we introduce a kernel function for trajectories inspired by STL robustness, enabling a faithful and interpretable mapping of input sequences into a symbolic embedding space. This allows classification and explanation to be jointly performed, with every decision grounded in logical, human-understandable concepts.
% , enabling classification and explainability within the same framework. We leverage a kernel-based embedding method to represent trajectories in terms of their similarity to predefined STL concepts, ensuring that explanations align with human-interpretable logic.
%This makes a neuro-symbolic time series classifier which is both accurate and interpretable, as every prediction is accompanied by a set of concepts, which are influential towards its characterisation. Such local explanations can be then combined to create class-descriptive formulae, i.e. global explanations for the outcome of the model. 
This yields (i) local explanations as human-readable STL conditions justifying individual predictions, and (ii) global explanations as class-characterising formulae. Experiments demonstrate that STELLE achieves competitive accuracy while providing logically faithful explanations, validated on diverse real-world benchmarks.
\end{abstract}

%% JAIR Note: 
%% Do not include ACM CCS Concepts or Keywords

%% To be updated by authors.
\received{October 2025}

\maketitle

\section{Introduction} \label{sec:intro}

Time series data are pervasive nowadays, arising from 
% technical 
systems such as industrial sensor readings, moving through our everyday life via data generated by Internet of Things devices, to safety-critical applications such as digital health equipment and autonomous vehicles. 
Being able to effectively and efficiently analyse time series is an objective of paramount importance, often tackled by means of data-driven Machine Learning (ML) algorithms \citep{Ruiz_Flynn_Large_Middlehurst_Bagnall_2021,dl-tsc}. 
Many of these models are however \textit{black-box}, i.e. the logic behind their outcomes is opaque, hindering their application in high-stakes scenarios, where safety and reliability need to be guaranteed. This lack of transparency raises ethical \citep{ethics1, ethics2} and legal \citep{legal, gdpr} concerns, and undermines human trust despite their state-of-the-art performance across various tasks \citep{LEICHTMANN2023107539}.
Time series, unlike images, are difficult to interpret by humans, having a hidden semantic that often needs domain knowledge to be grasped. 
Most existing Explainable Artificial Intelligence (XAI) frameworks for this data type are \textit{post-hoc} \citep{tsc-xai-survey}, i.e. explanations are added on top of a trained model and not computed alongside the predictions of the model itself, and many critical points have been raised about their effectiveness \citep{tsc-posthoc-survey}.

% STL (maybe example of property that can bexpressed in STL??)
A suitable formalism for describing time series behaviour and patterns is Signal Temporal Logic (STL) \citep{temporal-logic,stl}, which offers a rich yet concise language for expressing properties of signals varying over time. It avoids the vagueness and redundancy of natural language, while still being easy to translate in common words; indeed, STL is the de-facto standard language in Cyber-Physical Systems (CPS) applications \citep{stl-cps}. One task that received significant attention \citep{stl-mining-survey,roge,ir-bo} is the so-called \textit{requirement mining}, which given a set of (labelled) trajectories aims at extracting a set of STL properties able to discriminate or characterise them. The mined formulae, although exploitable as signal classifiers, only provide a global explanation for the learned decision boundary. 
% concept based as workaround 
Hence STL formulae are valuable candidates to serve as explanations in time series learning algorithms. Following this intuition, we propose a concept-based model \citep{ghorbani2019automatic, NEURIPS2020_ecb287ff} for time series classification, where concepts are represented as STL requirements. Differently from post-hoc XAI methods, our model is \textit{interpretable by design}, as explanations are computed concurrently with the predictions, in the form of succinct sets of STL conditions. Doing so, the outcome of our model is both human-interpretable and able to extract actionable knowledge from the input time series, since the learned concepts can be used to monitor future behaviour of the system or to compare predictions with previous knowledge. Furthermore, an interpretable approach can help meet regulatory requirements, ensuring that the deployment of these technologies adheres to legal standards and ethical guidelines.

% contributions 
Specifically, our contributions consist in: (i) extending the concept of the STL kernel proposed by \citep{stl-kernel} to embed trajectories into STL formulae; (ii) proposing STELLE\footnote[1]{Italian term for ``stars''}, a concept-based time series classification architecture that directly embeds raw trajectories into a symbolic space of STL formulae through a novel robustness-inspired kernel, enabling end-to-end interpretable classification grounded in temporal logic semantics; (iii) introducing a dual explanation framework that generates both local explanations (as STL formulae justifying individual predictions) and global explanations (as class-characterising temporal patterns) from the same concept space; (iv) demonstrating
% through extensive experiments 
that STELLE achieves competitive accuracy while maintaining interpretability across diverse real-world benchmarks, including high-dimensional time series data.

% powered by a novel kernel for trajectories, which embeds raw time series into a symbolic space structured by STL robustness, and leveraging kernel based methods for STL~\citep{stl-kernel} to enable interpretable and logic-grounded classification;
% % proposing a concept-based time series classification model. Our Neuro-Symbolic (NeSy) architecture embeds STL formulae into a shared semantic space leveraging kernel based methods for STL~\citep{stl-kernel}, enabling classification and explainability within the same framework; 
% (ii) providing both local and global concept-based explanations, which respectively allow us to interpret the outcome of the model for a specific input trajectory and the overall behaviour of the classifier; (iii) experimentally proving the effectiveness of our model on multiclass classification examples collected from various applications and its scalability on synthetic datasets of increasing dimension. 

The remainder of this paper is organised as follows: Section~\ref{sec:background} introduces the necessary background on concept-based learning, explanation by backpropagation, and Signal Temporal Logic and its kernel. Section~\ref{sec:temb} details proposes our extension of the STL kernel to embed trajectories. Section~\ref{sec:model} presents the STELLE framework, detailing its architecture and explanation generation process. In Section~\ref{sec:experiments}, the model is evaluated against state-of-the-art methods across both and its interpretability metrics are presented. Lastly, Section~\ref{sec:rw} discusses related work.

\section{Background} \label{sec:background}+
This section presents the theoretical foundations of our approach. We begin by reviewing concept-based models, which enable interpretable reasoning through human-understandable latent representations, and describe the backpropagation-based technique employed for explanation extraction. We then outline Signal Temporal Logic (STL), a formal language for specifying temporal properties of time series, and discuss its quantitative semantics. Finally, we describe a kernel formulation for STL that provides a similarity measure between logical specifications, forming the basis for the trajectory embedding kernel used in our architecture.
% \subsection{Concept-based models}
\subsection{Concept-based models}
Concept-based models are an emerging paradigm in machine learning that integrate human-understandable concepts into the decision-making process of the model, to make it inherently interpretable \citep{ghorbani2019automatic, NEURIPS2020_ecb287ff}.
Unlike post-hoc explainability frameworks, which attempt to rationalise black-box predictions after training, concept-based approaches embed semantic representations directly within the learning architecture. These concepts capture meaningful, domain-specific patterns that contribute explicitly to the downstream task.

In the context of time series classification, concepts can correspond to temporal or frequency-based phenomena, such as periodic trends, abrupt changes, local motifs, or statistical features (e.g., amplitude, duration, variability), that experts naturally use to interpret temporal data. By associating model decisions with such interpretable temporal concepts, these frameworks provide insights not only into what the model predicts, but also why certain temporal patterns are deemed discriminative. This paradigm thus bridges the gap between model transparency and performance, enabling more trustworthy and diagnostically useful time series models. %, as explored in recent works such as \citet{concepts1, concepts2}.

% Concepts should represent domain-specific knowledge, to foster comprehension of the predictions when provided as explanations; differently from post-hoc XAI frameworks, they are incorporated in the learning process and directly contribute to the downstream classification task. 
% For example, in image classification tasks concepts might represent the colour and/or the shape attributes of image subjects.
% There are mainly two classes of concept-based models: those requiring concepts to be provided as input at training time \citep{concept-bottleneck,barbiero-concept} and those automatically extracting concepts, without the need of any annotation \citep{ghorbani2019automatic,img1}.

\subsection{Explanation by backpropagation} \label{sec:ig}
A common family of explanation methods for neural networks is based on the backpropagation of relevance scores from the output layer to the input space.  
These methods aim to assign an attribution score to each input feature, reflecting its contribution to the model’s prediction.  
Formally, given a differentiable model $F: \mathbb{R}^d \to [0,1]^K$, $\mathbf x \mapsto \hat y$, $K$ number of classes, the relevance of input dimension $x_i$ for class $\hat{y}$ can be expressed as the partial derivative $\frac{\partial F_{\hat{y}}(\mathbf{x})}{\partial x_i}$, which quantifies local sensitivity.  
However, simple gradient-based explanations suffer from noise and saturation effects, providing unstable or uninformative attributions.

To mitigate these issues, \citet{integratedgradients} introduced \emph{Integrated Gradients} (IG), which compute attributions by integrating the model’s gradients along a continuous path from a baseline input $\mathbf{x}'$ (typically representing ``absence of signal'') to the actual input $\mathbf{x}$.  
Formally, for the target class $\hat{y}$, the IG attribution for the $i$-th feature is defined as:
\begin{equation*}
\mathrm{IG}_i(\mathbf{x}) = (x_i - x_i') \int_{0}^{1} 
\frac{\partial F_{\hat{y}}\big(\mathbf{x}' + \alpha (\mathbf{x} - \mathbf{x}')\big)}{\partial x_i} \, d\alpha
\end{equation*}
This integral captures the cumulative effect of feature $x_i$ on the class score $F_{\hat{y}}$ along the interpolation path between $\mathbf{x}'$ and $\mathbf{x}$. In this formulation, $\alpha \in [0, 1]$ is a continuous scaling parameter that traces the interpolation path between the baseline input $\mathbf{x}’$ and the actual input $\mathbf{x}$. When $\alpha = 0$, the model is evaluated at the baseline (typically representing the absence of signal), and when $\alpha = 1$, it is evaluated at the true input. Intermediate values of $\alpha$ correspond to points along the straight line connecting $\mathbf{x}’$ and $\mathbf{x}$ in feature space. The integral thus accumulates the gradient contributions of feature $x_i$ along this path, capturing how changes in $x_i$ influence the model’s output as the input transitions from baseline to actual data. \\
Integrated Gradients satisfy desirable properties such as \textit{sensitivity} (non-zero attribution for relevant inputs) and \textit{implementation invariance} (consistent results for functionally equivalent networks), making them a robust choice for model interpretability.  

In this work, we apply Integrated Gradients not to raw inputs but to the latent concept activations $\mathbf{z}(\mathbf{x})$, yielding class-specific weights $\mathbf{W}_{\hat{y}}$ that quantify the relevance of each concept to the final decision.

\subsection{Signal Temporal Logic (STL)} STL is a linear-time temporal logic which expresses properties on trajectories over dense time intervals \citep{stl}. We define as trajectories the functions $\tau: I\rightarrow \mathcal{X}$, where $I\subseteq \mathbb{R}_{\geq 0}$ is the time domain and $\mathcal{X}\subseteq \mathbb{R}^d$, where $d\in \mathbb{N}$ is the state space. The syntax of STL is given by:
\begin{equation*}
\varphi:=\top\mid\pi\mid\lnot\varphi\mid \varphi_1\land\varphi_2\mid\varphi_1\mathbf{U}_{[a, b]}\varphi_2
\end{equation*}
where $\top$ is the Boolean \emph{true} constant; $\pi$ is an \emph{atomic predicate}, interpreted as a function of the form $f_{\pi}(\mathbf{x})\geq 0$ over variables $\mathbf{x}\in \mathbb{R}^{N}$ (we refer to $N$ as the number of variables of a STL formula, i.e. individual signals used as arguments of the atomic predicates);
%$\lnot$ and $\land$ are the Boolean \emph{negation} and \emph{conjunction}, respectively (from which the \emph{disjunction} $\lor$ follows by De Morgan's law); $\mathbf{U}_{[a, b]}$, with $a, b \in \mathbb{Q}, a<b$, is the \emph{until} operator, from which the \emph{eventually} $\mathbf{F}_{[a, b]}$ and the \emph{always} $\mathbf{G}_{[a, b]}$ temporal operators can be deduced.
negation $\neg$ and conjunction $\land$ are the standard Boolean connectives, and $\mathbf{U}_I $ is the \textit{Until} temporal modality, where $[a,b]\subseteq I$ is a real positive interval. We can derive the disjunction operator $\lor$ from $\land$ and $\neg$ vie De Morgan's law as $\varphi_1 \lor \varphi_2 = \neg(\neg\varphi_1 \land \neg\varphi_2)$, the \textit{Eventually} $\mathbf{F}_{[a,b]} $ and \textit{Always} $\mathbf{G}_{[a,b]}$ operators from the Until temporal modality as $ \mathbf{F}_{[a,b]} \varphi = \top \, \mathbf{U}_{[a,b]} \varphi$ and $\mathbf{G}_{[a,b]} \varphi = \neg\mathbf{F}_{[a,b]} \neg \varphi$, respectively.

We can intuitively interpret the temporal operators over a time interval $[a, b]$ as:
\begin{itemize}
  \item \emph{Eventually}: a property is eventually satisfied if it is satisfied at some point inside $[a, b]$;
  \item \emph{Globally}: a property is globally satisfied if it is true continuously in the future in the specified temporal interval; 
  \item \emph{Until}: it captures the relationship between two conditions $\varphi, \psi$ in which the first condition $\varphi$ holds until, at some point in $[a, b]$ in the future, %the second condition 
  $\psi$ becomes true.
\end{itemize}
% a property is \textit{eventually} satisfied if it is satisfied at some point inside the temporal interval, while a property is \textit{globally} satisfied if it is true continuously in $[a, b]$; finally the \textit{until} operator captures the relationship between two conditions $\varphi, \psi$ in which the first condition $\varphi$ holds until, at some point in $[a, b]$, the second condition $\psi$ becomes true.
We call $\mathcal{P}$ the set of well-formed STL formulae; we refer to the number of symbols of an STL formula as its number of nodes, and to the maximum number of allowed signals dimension over which it can be interpreted as the number of its variables. 
For example, the sentence ``the temperature $\tau$ of the room will reach $25$ degrees within the next $10$ minutes and will stay above $22$ degrees for the following $60$ minutes'' translates in STL as $\mathbf{F}_{[0, 10]} (\tau \geq 25 \wedge \mathbf{G}_{[0, 60]} \tau \geq 22)$.
% STL is endowed with both a \emph{qualitative} (or Boolean) semantics, giving the classical notion of satisfaction of a property over a trajectory, i.e.\ $s(\varphi, \tau, t) = 1$ if the trajectory $\tau$ at time $t$ satisfies the STL formula $\varphi$, and a \emph{quantitative} semantics, denoted by $\rho(\varphi, \tau, t)$. The latter, also called \emph{robustness}, is a measure of how robust is the satisfaction of $\varphi$ w.r.t. perturbations of the signals. A formal definition of the syntax and semantics of STL is given in Appendix~\ref{app:stl}. 
\subsubsection{Robustness}
STL has both a qualitative (or Boolean) and a quantitative (or robust) semantics \cite{stlrob}. While the Boolean semantics gives the classical notion of satisfaction, the robustness of a formula $\varphi$ on a signal $\tau$ is a value $\rho(\varphi, \tau, t) \in
% $\bar{\mathbb{R}}$\footnote{$ \bar{\mathbb{R}} =
\mathbb{R} \cup \{ - \infty , + \infty\} $ quantifying the \textit{robustness degree} of the property $\varphi$ on the trajectory $\tau$ at time $t$. Intuitively, robustness generalises Boolean satisfaction by measuring not only whether a formula holds, but also how strongly it is satisfied or violated. Positive robustness values indicate that $\varphi$ is satisfied with a safety margin, whereas negative values quantify the extent of its violation. Its sign provides the link with the standard Boolean semantics, in that a signal $\tau$ satisfies an STL formula $\varphi$ at a time $t$ iff the robustness degree $\rho(\varphi,\tau,t) \ge 0$. Its absolute value, instead, can be interpreted as a measure of the distance to violation or satisfaction of $\varphi$ on $\tau$. 
% the robustness of the satisfaction with respect to noise in signal $\xi$, measured in terms of the maximal perturbation in the secondary signal $\theta(\xi(t))$, preserving truth value. 
Robustness is recursively defined as:
\begin{alignat*}{2}
 \rho(\pi,\tau,t) &= f_\pi(\tau(t)) \qquad \text{for } \pi(\mathbf{x})=\big(f_\pi(\mathbf{x})\geq 0\big)\\
 \rho(\lnot\varphi,\tau,t) &= -\rho(\varphi,\tau,t)\\
 \rho(\varphi_1\land\varphi_2,\tau,t) &= \min\big(\rho(\varphi_1,\tau,t), \rho(\varphi_2,\tau,t)\big)\\
 \rho(\varphi_1\mathbf{U}_{[a, b]}\varphi_2,\tau,t) &= \max_{{t'\in[t+a,t+b]}}\big(\min\big(\rho(\varphi_2,\tau,t'), \\ & \qquad \qquad \qquad
 \min_{t''\in[t,t']}\rho(\varphi_1,\tau,t'')\big)\big) 
\end{alignat*}
For completeness, we report also the definition of robustness of derived temporal operators: 
\begin{alignat*}{2}
\text{eventually} \quad  &\rho(\mathbf{F}_{[a, b]}\varphi, \tau, t)  &&= \max_{t'\in[t+a,t+b]} \rho(\varphi,\tau,t') \\
\text{always} \quad &\rho(\mathbf{G}_{[a, b]}\varphi, \tau, t) &&= \min_{t'\in[t+a,t+b]} \rho(\varphi,\tau,t')
\end{alignat*}

Throughout the paper, we will consider this metric at $t=0$ and simplify the notation to $\rho(\varphi, \tau)$. 
A distribution $\Gamma$ over STL formulae can be constructed procedurally by recursively expanding syntax trees, where each node has a fixed probability of being instantiated as an atomic predicate, and the remaining cases are uniformly sampled from the available temporal and logical operators \citep{stl2vec}.

\subsection{A kernel for STL} \label{sec:kstl}
A kernel for STL that leverages the quantitative semantics of STL \citep{stl-kernel} can be used for finding continuous representations of STL formulae. Indeed, robustness in this context allows formulae to be treated as functionals that map trajectories into real numbers i.e. ${\rho(\varphi,\cdot): \mathcal{T}\rightarrow \mathbb{R}}$ such that $\tau\mapsto \rho(\varphi, \tau)$. By considering these functionals as feature maps and fixing a probability measure $\mu_0$ on the trajectory space $\mathcal{T}$, a kernel function that captures the similarity between STL formulae $\varphi$ and $\psi$ can be defined as follows:
% \small
\begin{equation}
k(\varphi, \psi) = \langle \rho(\varphi, \cdot), \rho(\psi, \cdot) \rangle = \int_{\tau\in \mathcal{T}} \rho(\varphi, \tau) \rho(\psi, \tau) \, d\mu_0(\tau)
\label{eq:stl-kernel} % referenziata in appendice
\end{equation}
\normalsize
This approach enables the use of the scalar product in the Hilbert space $L^2$ as a kernel for $\mathcal{P}$. Essentially, this kernel yields a high positive value for formulae that behave similarly on high-probability trajectories w.r.t. $\mu_0$, and a low negative value for formulae that disagree on those trajectories. Further details about the STL kernel are provided in Appendix~\ref{app:kernel}.

This kernel establishes the mathematical foundation for our framework by enabling two key capabilities: (i) quantifying semantic similarity between temporal logic formulae through their behavioural profiles, and (ii) providing the theoretical bridge between symbolic reasoning and continuous embedding spaces that we exploit for trajectory-formula alignment. While the original formulation measures formula-formula similarity, we extend this core definition to develop our trajectory embedding mechanism, maintaining interpretability through shared semantic grounding in STL robustness.

\section{Trajectory embedding kernel} \label{sec:temb}
The first contribution of this work is the definition of a trajectory embedding kernel that enables the representation of raw time series data within the same interpretable space as logical formulae. This kernel constitutes a fundamental component of the proposed architecture, as it allows the model to jointly reason over trajectories and temporal logic concepts.

While the STL kernel described in Section~\ref{sec:kstl} enables the embedding of logical formulae, it is key to our goal to embed raw input trajectories into the same interpretable space. To achieve this, we define a trajectory-specific kernel inspired by STL robustness. \
A kernel for input trajectories can be defined in the same functional space as STL formulae by extending the concept of robustness to measure similarity between trajectories. 
Given a reference trajectory $\tau \in \mathcal{T}$, we define a function $\rho_\tau : \mathcal{T} \rightarrow \mathbb{R}$ that quantifies the similarity between $\tau$ and another trajectory $\xi \in \mathcal{T}$. A natural choice for this function is 
\begin{equation}    \label{eq:epsilon} % uso in appendice
\rho_\tau(\xi) = d_\tau(\xi)/\varepsilon \qquad \text{where} \qquad d_\tau(\xi) = \| \xi - \tau \|_2^2
\end{equation}
is the squared $L_2$ distance and $\varepsilon > 0$ controls the locality of the similarity measure.
% Building on the STL kernel framework, we develop a dual embedding mechanism that projects raw trajectories into the same interpretable space as temporal logic formulae. The key insight is to treat trajectories as functionals through a novel similarity measure $\rho_\tau : \mathcal{T} \rightarrow [-1, 1]$, defined for any $\xi, \tau \in \mathcal{T}$ as:
% \begin{equation} \label{eq:rho_tau}
% \rho_{\tau}(\xi) = 2\cdot \exp \bigg( - \frac{d_{\tau}(\xi)}{\varepsilon}\bigg)-1 % \quad \in [-1,1]    
% \end{equation}
% where $d_\tau(\xi)$ denotes a distance metric between trajectories, such as the squared $L_2$ norm $d_\tau(\xi) = || \xi - \tau ||_2^2$, and $\varepsilon > 0$ controls locality. 
% This construction ensures that $\rho_\tau$ achieves values close to $1$ when $\xi$ is similar to $\tau$, and values approaching $-1$ when the dissimilarity increases. \\
By interpreting $\rho_\tau$ as a trajectory-dependent functional analogous to the robustness of an STL formula, we can define a kernel between a trajectory $\tau$ and an STL formula $\varphi$ as:
\begin{equation}
k(\tau, \varphi) = \int_{\xi \in \mathcal{T}} \rho_\tau(\xi) \rho(\varphi, \xi) \, d\mu_0(\xi)
\label{eq:traj-formula-kernel}
\end{equation}
This formulation has three critical properties: (i) trajectories and formulae occupy a shared semantic space where distances reflect behavioural alignment, (ii) the kernel inherits the STL kernel's theoretical guarantees, and (iii) all subsequent operations remain grounded in temporal logic semantics. Section~\ref{sec:architecture} depicts how STELLE exploits this through direct concept-based embedding, avoiding intermediate representations while preserving interpretability.

% This formulation enables embedding trajectories into the same Hilbert space $L^2$ used for STL formulae, allowing joint comparison and integration of logic-based representations within a unified framework. For this reason, this kernel forms the basis of our trajectory embedding mechanism in the classification architecture (see Section~\ref{sec:architecture}), ensuring that the input is represented in terms of its behavioural alignment with interpretable logic-based concepts. Further details about the kernel are provided in Appendix~\ref{app:kernel}.

\section{The model} \label{sec:model}
We now introduce STELLE (Signal Temporal logic Embedding for Logically-grounded Learning and Explanation), a novel architecture for interpretable time series classification. STELLE operates by projecting input trajectories into a semantic space defined by a diverse set of human-interpretable Signal Temporal Logic (STL) concepts. The model learns to classify by attending to these concepts and assessing how unusually they are satisfied relative to different classes. This design inherently provides explanations: local predictions are justified by a concise logical combination of relevant temporal patterns, while global class characterisations are derived by aggregating these local explanations into discriminative formulae. The following subsections detail the construction of the concept set, the STELLE architecture, and the procedures for extracting both local and global explanations.

\subsection{Concept set}  \label{sec:cset}
% come tiro fuori set di concetti
In our model, concepts are defined as temporal logic conditions expressed using STL formulae, which encapsulate human-interpretable patterns over time. These logical expressions serve both as interpretable units of explanation and as semantic anchors in the shared embedding space.

To build a semantically rich yet interpretable concept set, we generate STL formulae by sampling from a space of symbolic templates. Each formula is instantiated from a probabilistic grammar with a bounded number of variables and a maximum number of nodes, ensuring syntactic simplicity and human interpretability. These candidate formulae are evaluated over a fixed dataset of time series signals using quantitative semantics, resulting in robustness signature vectors that capture their behaviour across trajectories.
\\Concept selection proceeds incrementally: at each step a batch of new candidates is sampled and evaluated, and those whose signatures differ significantly from the existing set are retained, measured via cosine similarity. To promote parsimony, structurally simpler formulae are preferred when behavioural redundancy is detected among candidates. This process continues until the desired number of concepts is reached. 
Importantly, a concept set is generated once using a fixed temporal window, and the resulting STL formulae can later be linearly rescaled to accommodate datasets of varying lengths by adjusting their temporal thresholds proportionally.

This process guarantees that the selected formulae are both representative and diverse, and that they span a broad range of temporal patterns observed in the data. We refer the reader to Appendix~\ref{app:concept_selection} for a more detailed explanation of the procedure.

\subsection{Architecture}
\begin{figure}
    \centering
    \includegraphics[width=0.8\linewidth]{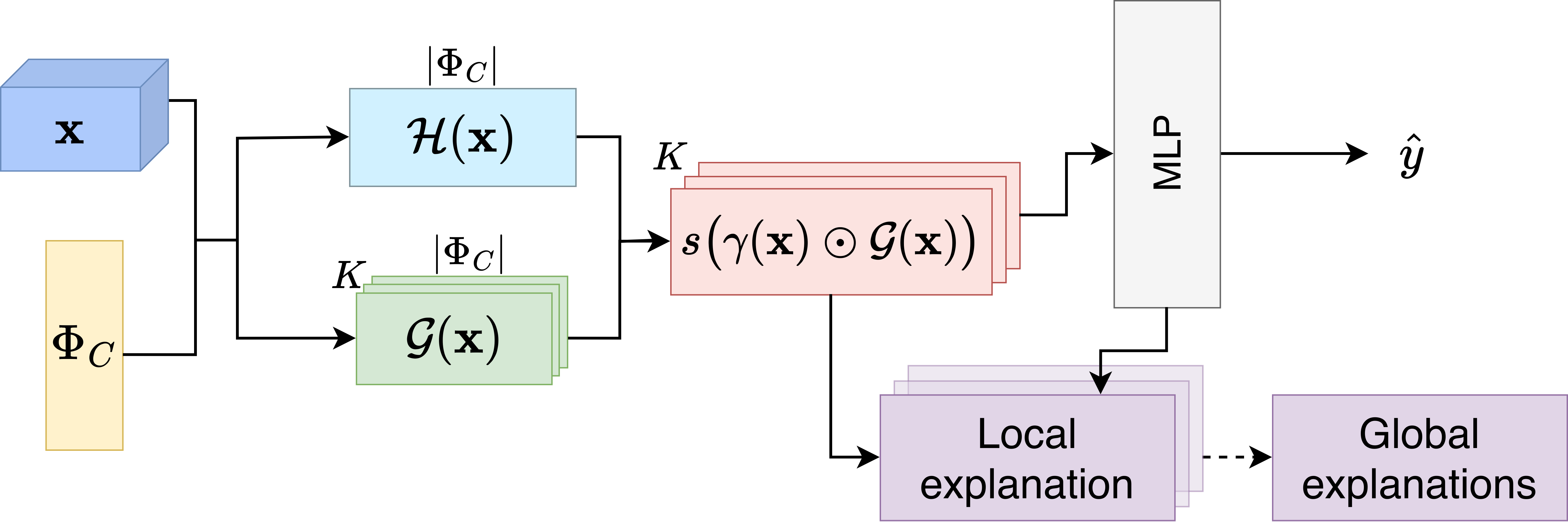}
    \caption{STELLE architecture. Input trajectories $\mathbf{x}$ are embedded via STL concepts formulae $\Phi_C$, creating $\mathcal{H}(\mathbf{x})$. This gets scaled, creating $\gamma(\mathbf{x})$, and integrated with class-specific scores $\mathcal{G}(\mathbf{x})$ for classification $\hat y$ and generation of local explanations, which can be aggregated into global explanations.}
    \label{fig:architecture}
\end{figure}

Our architecture, STELLE, % (Signal Temporal logic Embedding for Logically-grounded Learning and Explanation), 
is designed to perform interpretable time series classification through concept-based reasoning. It integrates symbolic information from Signal Temporal Logic (STL) with trajectory representations, enabling explanations that are grounded in logical temporal behaviours. The architecture, depicted in Figure~\ref{fig:architecture}, consists of three components: 
%(i) concept embedding, 
(i) a trajectory embedding, (ii) a discriminability mechanism, and (iii) a learned concept relevance module. %a soft attention-based classification module.

The model supports an arbitrary number of classes, enabling it to scale naturally to multiclass classification tasks, and each formula may involve one or more signal dimensions, allowing the model to express and attend to complex relationships in multivariate inputs.

% \paragraph{Concept embedding}
% % To enable symbolic reasoning over interpretable logical properties, we introduce a set of $n$ STL formulae ${\varphi_1, \dots, \varphi_n}$, which act as high-level concepts for the model to attend to. Each of these formulae is embedded into the same space as the input trajectories using the STL kernel introduced in Section~\ref{sec:background}, yielding a concept embedding matrix $\text{cemb} \in \mathbb{R}^{n \times m}$. Each row of $\text{cemb}$ corresponds to a concept prototype, expressed as a point in the base embedding space defined by the anchor set ${\psi_1, \dots, \psi_m}$.
% We begin by defining a fixed set of $n$ STL formulae $\Phi_C = \{\varphi_1, \dots, \varphi_n\}$ that represent high-level interpretable concepts, using the procedure described in Section~\ref{sec:cset}. These formulae are embedded into a symbolic space via the STL kernel introduced in Section~\ref{sec:background}, resulting in a matrix $\mathcal{H}_{\Phi_C} \in \mathbb{R}^{n \times m}$, where $m$ is the number of anchor formulae $\Phi_A = \{\psi_1, \dots, \psi_m\}$. Each row of $\mathcal{H}_{\Phi_C}$ defines a concept prototype as a point in the embedding space structured around these anchors. This ensures that all concepts lie in a space where dimensions correspond to logical properties and distances reflect behavioural similarity.

\paragraph{Trajectory embedding}
A key innovation of our architecture lies in the use of a robustness-inspired kernel for input trajectories introduced in Section~\ref{sec:temb}, which enables their projection into the same logic-grounded space as STL formulae.
% To map input time series into the shared symbolic space, we employ the robustness-based kernel introduced in Section~\ref{sec:temb}. This kernel quantifies the behavioural similarity between a trajectory and a set of STL formulae, enabling a semantically meaningful comparison between raw input data and logical concepts. \\
% Formally, given an input time series $\mathbf{x}$, we compute its embedding $\mathcal{H} ({\mathbf{x}}) \in \mathbb{R}^{m}$ by evaluating its similarity to a fixed set $\Phi_A$ of $m$ anchor formulae. The resulting vector encodes how well $\mathbf{x}$ satisfies each of the anchor formulae, producing an interpretable representation aligned with human-understandable semantics.

We begin by defining a fixed set of STL formulae $\Phi_C = \{\varphi_1, \dots, \varphi_{C}\}$ that represent high-level, human-interpretable concepts, as described in Section~\ref{sec:cset}. Given an input time series $\mathbf{x}$, we compute its embedding $\mathcal{H} ({\mathbf{x}}) \in \mathbb{R}^{C}$ by evaluating the robustness of $\mathbf{x}$ with respect to each concept formula $\varphi_i \in \Phi_C$, as described in Section~\ref{sec:temb}. This results in a concept-based representation of the trajectory, where each dimension quantifies the degree to which the corresponding temporal property is satisfied. The resulting embedding is fully interpretable, as each component is tied to a known STL formula.

\paragraph{Discriminability scores}
To assess the explanatory value of each concept, we compute class-specific discriminability scores based on robustness statistics. Let $K$ be the number of classes, and let $k$ denote a specific class. For each concept $\varphi_i \in \Phi_C$, we measure how atypical its satisfaction is for the input trajectory when compared to other classes. We define the discriminability matrix $\mathcal{G}({\mathbf{x}}) \in [0, \infty)^{C \times K}$, where each entry $\mathcal{G}_i^k (\mathbf{x} )$ is computed as
\begin{equation*}
 \mathcal{G}_i^{k}( \mathbf{x} ) = \frac{|\rho(\varphi_i, \mathbf{x}) - \mu_{k^*, i}|}{\sigma_{k^*, i} + \varepsilon_G}
 \end{equation*}
Here, $\rho(\varphi_i, \mathbf{x})$ denotes the robustness of the input trajectory $\mathbf{x}$ with respect to formula $\varphi_i$, $\varepsilon_G$ is a small regularisation constant, and $\mu_{k^*, i}$, $\sigma_{k^*, i}$ are the mean and standard deviation of robustness values for class $k^* \ne k$. This normalisation highlights the extent to which each formula supports or contradicts the model’s prediction relative to the distribution of robustness in the competing classes.

% By exploiting the same $n$ STL formulae $\{\varphi_1, \dots, \varphi_n\}$ introduced above, we compute a tensor of discriminability scores $\text{G}(x) \in \mathbb{R}^{n \times c}$, where $c$ is the number of classes. \\This representation reflects the per-class deviation of robustness values from typical behaviour observed in the training data. For each class $l \in \{0, 1\}$, we define:
%  
%     \begin{equation}
%         G_l(x)_i = \frac{r_i(x) - \mu_{\hat l, i}}{\sigma_{\hat l, i}}, \quad i = 1, \dots, n
%     \end{equation}
%   
% where $r_i(x)$ is the raw robustness of $x$ with respect to formula $\varphi_i$, and $\mu_{\hat l, i}$, $\sigma_{\hat l, i}$ denote the mean and standard deviation of robustness for formula $\varphi_i$ computed on class $\hat{l}$, the opposite of the predicted class of $x$. This normalisation highlights how strongly a formula supports or contradicts a classification, based on how unusual the behaviour is for the opposing class.

\paragraph{Learned concept relevance} \label{sec:architecture} % or Concept fusion mechanism
% To direct the focus on the most relevant concepts, we apply a soft attention mechanism between the trajectory embedding $H_{\mathbf{x}}$ and the concept embeddings $H_{\Phi_C}$. We use a scaled dot-product attention:
To identify the most relevant concepts for each input, we compute concept selection scores that determine their relative importance. Specifically, we generate raw selection weights as:
\begin{equation} \label{eq:T}
\gamma(\mathbf{x}) = \mathcal{H}({\mathbf{x}}) / T \in \mathbb{R}^{C}
\end{equation}
where $T$ is a temperature parameter controlling the sharpness of the selection.
We standardise the trajectory embeddings $\mathcal{H}({\mathbf{x}})$ before computing $\gamma(\mathbf{x})$ to ensure stable and comparable concept weighting across samples. This normalisation step prevents scale disparities among embeddings from dominating the relevance computation. In contrast, $\mathcal{G}(\mathbf{x})$ is left unstandardised, as its values are already expressed in a relative scale that captures concept distinctiveness across classes, and rescaling it would distort its interpretive meaning.
\\These selection weights are then used to modulate the class-specific discriminability scores through an element-wise normalisation operation, producing the combined representation:
% \begin{equation}\label{eq:z}
% \mathbf{z}(\mathbf{x}) = s\big(\gamma(\mathbf{x})\odot \mathcal{G}({\mathbf{x}})\big) \quad \in (-1,1)^{{C}\times K}
% \end{equation}
\begin{alignat}{2}
\mathbf{z}(\mathbf{x}) &= \gamma(\mathbf{x})\odot \mathcal{G}({\mathbf{x}}) &\quad&\in [0, \infty)^{{C}\times K} \label{eq:z}\\
\mathbf{\bar z}(\mathbf{x}) &= s(\mathbf{z}(\mathbf{x})) &&\in (-1,1)^{{C}\times K} \nonumber
\end{alignat}
where $\odot$ is the Hadamard product $s$ is the Softsign activation function.
This formulation ensures that the modulated concepts reflect both their relative relevance to the input and their class-specific explanatory strength.
\\The resulting matrix $\mathbf{\bar z}(\mathbf{x})$ is flattened and passed through an MLP to produce the final classification logits.

% The weights $\alpha(\mathbf{x})$ indicate the relative relevance of each concept in the classification of $\mathbf{x}$. These weights are used to modulate the discriminability scores, producing the matrix
% \begin{equation} \label{eq:z}
% \mathbf{z}(\mathbf{x}) = \alpha(\mathbf{x}) \odot \mathcal{G}({\mathbf{x}}) \in \mathbb{R}^{n \times K}   
% \end{equation}
% which combines concept relevance and class-specific explanatory strength. 
% This matrix is then flattened and passed through an MLP to produce the logits.
% \begin{equation*}
% \hat{y} = \mathbf{W} \cdot \text{flatten}(\mathbf{z}) + \mathbf{b}
% \end{equation*}
% with $\mathbf{W} \in \mathbb{R}^{K \times nK}$ and $\mathbf{b} \in \mathbb{R}^K$ as learnable parameters. \\

This architecture achieves interpretability in two complementary ways. First, the learned relevance weights $\gamma(\mathbf{x})$ identify which STL concepts contribute most strongly to the decision. Second, the discriminability scores $\mathcal{G}(\mathbf{x})$ explain why those concepts support or contradict the model’s prediction, by quantifying how unusual their satisfaction is for the predicted class compared to others. Combined, these two components provide both concept-level attribution and class-level reasoning, enabling explanations that are both local (specific to individual trajectories) and global (aggregated across the dataset). This dual interpretability allows STELLE to not only highlight the temporal patterns driving a particular prediction but also to characterise the broader logic underlying each class.

\subsection{Explanation extraction} \label{sec:explextraction}
The explanation generation synthesises information from three components: 
(i) the selection weights $\gamma(\mathbf{x})$ identifying relevant concepts, 
(ii) the discriminability scores $\mathcal{G}({\mathbf{x}})$ measuring concept atypicality, and 
(iii) the backpropagated classifier weights $\mathbf{W}$ determining each concept's importance to the final prediction.

\subsubsection{Local explanations}  
For a given input trajectory $\mathbf{x}$ and a target class $k$, local explanations quantify the contribution of each concept $\varphi_i$ to the model’s confidence in class $k$.  
We first compute the class-specific latent representation of $\mathbf{z}(\mathbf{x})$ as defined in Equation~\ref{eq:z}, where $\mathbf{z}(\mathbf{x})_k \in [0, \infty)^C$ denotes the vector of concept activations for $\mathbf{x}$ wrt class $k$, and $C$ is the total number of STL concepts. The index $k \in {1, \ldots, K}$ selects the subspace corresponding to the target class.

To estimate how each concept influences the class score, we employ \emph{Integrated Gradients} (see Section~\ref{sec:ig}) to backpropagate relevance from the classifier output to the latent concept dimensions.  
Let $F: \mathbb{R}^C \to [0,1]^K$ denote the classification head mapping latent concept activations to class probabilities.  
For class $k$, the relevance of each latent input dimension $z_i$ is computed as:
\begin{equation*}
W_{k,i} = (z_i - z_i') \int_{0}^{1} 
\frac{\partial F_{k}\big(\mathbf{z}' + \alpha (\mathbf{z} - \mathbf{z}')\big)}{\partial z_i} \, d\alpha
\end{equation*}
where $\mathbf{z}'$ is a baseline latent input (set to the zero vector), and $F_{k}$ is the classifier output corresponding to class $k$.  
The resulting relevance weights $\mathbf{W}_{k}$ are normalised to $[0,1]$, providing a smooth and faithful attribution of each concept’s influence on $F_{k}$.  

Importantly, the attributions are backpropagated to the \emph{non-normalised} matrix $\mathbf{z}(\mathbf{x})$, rather than its soft-signed version $\bar{\mathbf{z}}(\mathbf{x})$. This ensures that gradient propagation is not distorted by the bounded Softsign activation, which would otherwise attenuate large-magnitude responses and compress their relative importance. By operating directly on $\mathbf{z}(\mathbf{x})$, we preserve the true scale of the underlying concept activations, yielding explanations that more accurately reflect their quantitative influence on the prediction.

The final explanation vector is computed using a discriminative scoring mechanism that identifies concepts most distinctive to each class. For an input trajectory $\mathbf x$ and target class $k$, the concept-class activation matrix $\mathbf A (\mathbf x) \in \mathbb{R}^{C \times K}$, where $C$ is the cardinality of the concept set and $K$ is the number of classes, is defined as
\begin{equation*}
    \mathbf{r}(\mathbf{x})_{k} = \left| \mathbf{A}_{:,k}(\mathbf{x}) - \frac{1}{K-1} \sum_{^k* \neq k} \mathbf{A}_{:,k^*}(\mathbf{x}) \right|
\end{equation*}
where $\mathbf{A}_{:,k}(\mathbf{x})$ denotes the vector of concept activations for the predicted class $k$, and the summation term represents the mean activation vector across all other classes.
This formulation identifies concepts that exhibit substantially different activation patterns between the target class and alternative classes, thereby highlighting the most discriminative concepts for each classification decision and yielding interpretable, concept-level local explanations.

To identify the most influential concepts, two selection strategies are supported. If a fixed budget $\bar \gamma \in \mathbb{N}$ is specified, the top-$\bar \gamma$ concepts with highest absolute relevance scores $|\mathbf{r}|$ are selected. Otherwise, we apply a cumulative relevance cut-off: the entries of $|\mathbf{r}|$ are sorted, and the smallest prefix whose cumulative sum reaches a target fraction $\bar \gamma_t \in (0, 1]$ of the total relevance is retained. This allows the explanation size to adapt to the complexity of the input.

Let $\mathcal{F}_{\mathbf{x}} = \{\varphi_{1}, \dots, \varphi_{|\mathcal{F}_{\mathbf{x}}|}\}$ be the selected concepts. Each $\varphi_i$ is subsequently refined to enhance interpretability by adjusting its thresholds, polarity, and syntactic complexity, as will be detailed in Section~\ref{sec:simple}. Moreover, a threshold $ \bar \gamma_\ell \in [0, 1)$ controls the trade-off between formula complexity (node count) and classification fidelity: lower values prioritise classification accuracy over interpretability, allowing longer formulae that marginally improve performance. The result is a set of simplified formulae $\mathcal{F}'_{\mathbf{x}}$.
We define the local explanation for input $\mathbf{x}$ with reference to target class $\hat y$ as:
\begin{equation*}
E_\ell(\mathbf{x}, \hat y) = \bigwedge_{\varphi_i' \in \mathcal{F}'_{\mathbf{x}}} \varphi_i' % \varphi’_{1} \land \cdots \land \varphi’_{\gamma}, \quad \varphi’_{i} \in \mathcal{F}’
\end{equation*}

This logical expression provides a sufficient condition for describing class $\hat{y}$. Each conjunct represents a distinct, human-readable behavioural property. Together, they provide a faithful and compact explanation of the model’s decision.
Finally, the resulting formula goes through a logic-aware and data-aware pruning to remove uninformative predicates and improve its readability. We refer the reader to Appendix~\ref{app:phisimple} for more details about the process.

\subsubsection{Global explanations}
Global explanations aim to characterise, for each class $k$, the symbolic conditions that best describe its underlying temporal behaviour.  
Rather than aggregating only the explanations from correctly predicted test samples, we construct class-level formulae using the training trajectories and their true labels. This choice ensures that global explanations reflect intrinsic class semantics rather than potential model-specific biases introduced during prediction.  
In addition, since robustness evaluations are precomputed for all training trajectories, the explanation set can be efficiently updated when new data become available without retraining the full model.

For each class $k$, we collect all local explanations associated with the corresponding training trajectories:
\begin{equation*}
\mathcal{F}_k = \{E_\ell(\mathbf{x}, k) \ | \ \mathbf{x} \in \mathcal{T}_{\text{train}}, \ y(\mathbf{x}) = k \},
\end{equation*}
where $E_\ell(\mathbf{x}, k)$ denotes the local explanation extracted for class $k$ given input $\mathbf{x}$, $\mathcal{T}_{\text{train}}$ is the training set and $y: \mathcal{T}_{\text{train}} \to \mathbb{N}$, $y(\mathbf{x})$ is the true class of $\mathbf{x}$.  
Each local explanation $E_\ell(\mathbf{x}, k)$ consists of one STL formula $\varphi_i$ whose robustness values are most discriminative for that sample.

To identify which of these formulae exhibit general class-level relevance, we construct a division matrix $D^k \in \{0,1\}^{|\mathcal{F}_k| \times |\mathcal{F}_k|}$, where:
\begin{equation*}
D^k_{ij} = 
\begin{cases}
1, & \text{if } \rho(\varphi_j, \mathbf{x}_i) \ \notin [\min_{\mathbf{x}\notin y^{-1}{(k)}}\rho(\varphi_j, \mathbf{x}),\ \max_{\mathbf{x}\notin y^{-1}{(k)}}\rho(\varphi_j, \mathbf{x})] \\
0, & \text{otherwise}
\end{cases}
\end{equation*}
and $y^{-1}{(k)} = \{ \mathbf{x} \in \mathcal{T}_{train}\mid y(\mathbf{x}) = k\}$.\\
This criterion ensures that a formula $\varphi_j$ is considered discriminative for class $k$ if its robustness values on class-$k$ trajectories lie outside the range observed in all non-class-$k$ samples.  
The division matrix therefore quantifies, across all training data, which formulae distinctly separate the target class from the others.

Next, we seek a minimal subset of formulae $\mathcal{F}'_k \subseteq \mathcal{F}_k$ that jointly covers (i.e., distinguishes) all or most class-$k$ samples.  
To formalise this, we define a minimum-cost set cover problem:
\begin{equation*}
\min_{\mathbf c} \  \sum_i c_i \cdot \text{cost}(\varphi_i) \quad
\text{s.t.} \quad  D^k \mathbf{c} \geq \bar\gamma_g \mathbf{1}, \quad c_i \in \{0,1\},
\end{equation*}
where $\mathbf{c}$ is a binary selection vector indicating which formulae are included in the final global explanation, $\text{cost}(\varphi_i)$ corresponds to the syntactic complexity (e.g., node count) of $\varphi_i$, and $\bar \gamma_g \in [0,1)$ relaxes the coverage constraint to allow simpler, approximate global explanations.  
This optimisation is cast as a 0–1 Integer Linear Program (ILP). When the ILP becomes infeasible or too computationally expensive, we revert to a combinatorial search that greedily selects formulae whose disjunction maximises class separability while minimising redundancy.

The resulting symbolic explanation for class $k$ is expressed as:
\begin{equation*}
E_g(k) = \bigvee_{\varphi_i \in \mathcal{F}'_k} \varphi_i,
\end{equation*}
representing a sufficient logical condition for class membership, i.e. any trajectory satisfying at least one $\varphi_i$ in $\mathcal{F}'_k$ is typically assigned to class $k$.  

Crucially, this construction allows the global explanations to be incrementally refined: as new trajectories become available, robustness statistics and coverage relations can be updated without retraining, ensuring that $E_g(k)$ remains representative of the evolving data distribution.  
As in the local case, the final selected formulae undergo logic-aware and data-aware simplification to maximise readability while preserving discriminative and semantic consistency.

\subsubsection{Improving readability} \label{sec:simple}
To improve the readability and discriminative utility of learned STL formulae, we apply three complementary post-processing strategies. First, logical simplification rewrites formulae by eliminating redundant or structurally equivalent expressions based on syntactic rules (e.g., removing double negations, flattening nested temporal operators). Second, we apply threshold adjustment and ensure consistent satisfaction by negating the formula if it does not hold for the trajectories to explain, exploiting the linearity of robustness with respect to atomic thresholds to refine the decision boundary. By tuning thresholds and optionally negating formulae, we ensure that the resulting expression better separates the target and maintains interpretability. Lastly, data-aware simplification uses empirical evaluations over the trajectories to simplify subformulae that are uniformly true or false, enabling semantic reduction beyond structural rules. \\
For example, the formula 
$$\lnot (\mathbf G_{[0,20]}(\mathbf G_{[5,10]} \ (x_0 \leq 0.3))) $$
is progressively refined to
\begin{align*}
\lnot (\mathbf G_{[5,30]} \ (x_0 \le 0.3)) \\
\mathbf F_{[5,30]} \ (x_0 > 0.3)  
\end{align*}
through logical simplification.\\
Each method is independently applicable and contributes to enhancing clarity, compactness, and class-specific relevance. Full details rules and algorithms are presented in Appendix~\ref{app:phisimple}.

\section{Experimental evaluation} \label{sec:experiments}
We evaluate STELLE on a subset of multivariate time series classification tasks from the University of East Anglia (UEA) Time Series Archive \citep{bagnall2018ueamultivariatetimeseries}, a comprehensive benchmark suite spanning a wide range of domains, including physiological measurements, motion capture sequences, environmental sensor readings, and spectrographic signals. To construct a representative yet computationally tractable benchmark, we selected the first ten datasets in alphabetical order from the archive, excluding \textit{DuckDuckGeese}, which was omitted due to its extremely high dimensionality (over $1000$ variables) exceeding the available computational resources.
\\Table~\ref{tab:datasets} summarises the key characteristics of the selected datasets, including their short identifiers, size, sequence length, number of classes, and signal dimensionality. The \textit{Type} column indicates the application domain of each dataset: MOTION (motion capture), ECG (electrocardiography), HAR (Human Activity Recognition), SPECTRO (spectrographic signals) and EEG (electroencephalography).

\begin{table}
  \caption{Summary of the multivariate time series datasets used in the experimental evaluation. Each dataset is identified by a short two-letter code (ID). The dataset type denotes the domain of the recorded signals.}
  \label{tab:datasets}
\begin{tabular}{@{}lccccccc@{}}
\toprule

Dataset                   & ID & TrainSize & TestSize & Length & \#Classes & Type    & Channels \\ 
\midrule
ArticularyWordRecognition & AW & 275       & 300      & 144    & 25            & MOTION  & 9        \\
AtrialFibrillation        & AF & 15        & 15       & 640    & 3             & ECG     & 2        \\
BasicMotions              & BM & 40        & 40       & 100    & 4             & HAR     & 6        \\
Cricket                   & CR & 108       & 72       & 1197   & 12            & HAR     & 6        \\
ERing                     & ER & 30        & 270      & 65     & 6             & HAR     & 4        \\
Epilepsy                  & EP & 137       & 138      & 207    & 4             & HAR     & 3        \\
EthanolConcentration      & EC & 261       & 263      & 1751   & 4             & SPECTRO & 3        \\
HandMovementDirection     & HD & 160       & 74       & 400    & 4             & EEG     & 10       \\
Handwriting               & HW & 150       & 850      & 152    & 26            & HAR     & 3        \\
Libras                    & LI & 180       & 180      & 45     & 15            & HAR     & 2        \\  \bottomrule
\end{tabular}
\end{table}

After a first hyperparameter tuning performed on a few representative datasets, a number of hyperparameters were fixed, including batch size, the initialisation of $\varepsilon_G$, and the learning rates for both the latter and the temperature parameter $T$. Moreover, we fixed $\bar \gamma_t$ for explanation extraction and all the parameters regarding concept creation. For the remaining ones, due to the highly diverse nature of the datasets, we chose to fine-tune them tailored to each problem.
% Experiments were run with GPU allocations of up to 20GB and CPU allocations of up to 150GB (typically 50GB), depending on the size of the dataset.
\\For more information about training details, we refer the reader to Appendix~\ref{app:training}. All code and scripts for reproducing our experiments are available at \url{https://github.com/ireneferfo/STELLE}.

\paragraph{Baselines} 
To contextualise STELLE's performance, we compare it against a broad selection of state-of-the-art TSC methods. These include deep learning architectures such as InceptionTime \citep{inceptiontime}, ResNet \citep{dl-tsc}, TapNet \citep{tapnet}, and the HIVE-COTE 2.0 components HC1 \citep{Bagnall_2020} and HC2 \citep{Middlehurst_2021_HC}. We also consider transform-based methods including ROCKET \citep{rocket}, Arsenal and DrCIF \citep{Middlehurst_2021_HC}, as well as distance-based approaches using dynamic time warping (1NN-DTW-D, 1NN-DTW-I, 1NN-DTW-A \citep{dtw}). The comparison also includes symbolic and feature-based methods such as TSF \citep{tsf}, CIF \citep{Middlehurst_2020}, TDE \citep{TDE}, STC \citep{stc}, MrSEQL \citep{mrseql}, MUSE \citep{muse}, cBOSS \citep{cboss}, and RSF \citep{rsf}, which offer partial interpretability through engineered symbolic or interval features, though without explicit semantic explanations. We use performance metrics reported in \citet{Ruiz_Flynn_Large_Middlehurst_Bagnall_2021}, which represents the most comprehensive recent benchmark on the UEA multivariate archive.

\paragraph{Evaluation metrics}  
To comprehensively evaluate both predictive performance and interpretability, we employ a set of quantitative metrics spanning classification, explanation efficiency and readability, and computational efficiency.  

%For predictive performance, we report \textit{accuracy}, \textit{weighted accuracy} (to account for class imbalance), as well as \textit{sensitivity} and \textit{specificity}, which jointly characterise the model’s behaviour across different types of prediction errors.  

For explanation evaluation, we analyse both \textit{local} and \textit{global separability}. For each local explanation $E_\ell(\mathbf{x}, k)$, separability quantifies how well the explanation distinguishes between the target trajectory’s class and all other classes. Specifically, for a target trajectory $\mathbf{x}$ we compute:
\begin{equation} \label{eq:sepl}
\operatorname{Sep}(E_\ell(\mathbf{x}, k)) = 
\frac{\# \{y(\tau) \notin k : \operatorname{sign}(\rho(E_\ell(\mathbf{x}, k), \tau)) \neq \operatorname{sign}(\rho(E_\ell(\mathbf{x}, k), \mathbf{x}))\}}
{\# \{y(\tau) \notin k\}} \times 100,
\end{equation}
where $E_\ell(\mathbf{x}, k)$ denotes the local explanation extracted for class $k$, $\tau$ represents trajectories from the training dataset $\mathcal{T}_{train}$, and $y: \mathcal{T} \to \mathbb{N}$ maps trajectories to the target label.
This represents the percentage of trajectories from the opposite class whose robustness sign differs from that of the target trajectory, where $\tau$ denotes a trajectory from the set of all trajectories not belonging to class $k$.  
A high separability value indicates that the explanation generalises well to unseen samples of the same class and excludes those from other classes.  

We report the mean standard deviation of local separability under different filtering conditions:  
(i) only for correctly classified samples (reflecting explanation reliability),  
(ii) only for misclassified samples, analysed from two distinct perspectives: explanations generated with respect to the true class and with respect to the predicted class.
Explanations generated with respect to the true class reveal whether the model's reasoning aligns with actual class characteristics, measuring how well the explanation captures the true class's distinguishing features. Low separability here indicates a misalignment between the model's reasoning and ground truth. On the other hand, explanations generated with respect to the predicted class expose the flawed logic used to justify incorrect decisions, revealing what misleading patterns or correlations the model relies on when making errors. Analysing both perspectives provides comprehensive insight into explanation failure modes and model robustness.

For global explanations, we compute class separability both per-class and overall. The \textit{per-class separability} for class $k$ is given by:
\begin{equation*}
\operatorname{Sep}_k(E_g(k)) = \frac{TP_k + TN_k}{\text{total}_k},
\end{equation*}
where $TP_k$ and $TN_k$ respectively denote trajectories of class $k$ that satisfy the class explanation $E_g(k)$ and trajectories of other classes that do not satisfy it.

The \textit{overall separability} is obtained by micro-averaging across all classes:
\begin{equation} \label{eq:sep}
\operatorname{Sep} = \frac{\sum_{k} (TP_k + TN_k)}{\sum_{k} \text{total}_k},
\end{equation}
representing the total proportion of trajectories correctly classified by their respective class explanations.

We evaluate both measures under three conditions: the entire test set, only correctly classified samples, and only misclassified samples. This allows us to analyse how consistently the explanations reflect the model's decisions.

Finally, to assess interpretability and efficiency, we measure \textit{readability} (in terms of the number of syntactic nodes and distinct variables per formula), training time, inference time, and explanation extraction time.  
Together, these metrics provide a balanced assessment of both model performance and the interpretability-efficiency trade-off.

\subsection{Results} \label{sec:results}
% \usepackage{booktabs}
\begin{table}
% \centering
\caption{
Comparative performance analysis of STELLE against state-of-the-art time series classification methods on the UEA multivariate archive. The table presents classification accuracy scores with best results emphasised for easy comparison. The column ``Mean'' indicates the average performance of all baseline classifiers for each dataset, providing a reference for comparative ranking. Abbreviations used are: DTW-D (\textit{1NN-DTW-D}), DTW-I (\textit{1NN-DTW-I}), and DTW-A (\textit{1NN-DTW-A}). Performance metrics sourced from \citet{Ruiz_Flynn_Large_Middlehurst_Bagnall_2021}.}
\label{tab:smallresults}
\setlength{\tabcolsep}{4.3pt} %tight fit
% \usepackage{booktabs}
% \usepackage{booktabs}
\begin{tabular}{@{}lcccccccccc>{\columncolor[HTML]{dcdcdc}[1pt][1pt]}c@{}} 
\toprule
Dataset & DTW-D    & ROCKET       & STC           & TapNet & HC2           & DrCIF        & DTW-I & Arsenal      & RISE         & MrSEQL       & \textbf{STELLE} \\ \specialrule{0.07em}{0pt}{0pt} % \midrule
AW      & 0.99         & \textbf{1.0} & 0.98          & 0.97   & \textbf{1.0}  & 0.98         & 0.94      & \textbf{1.0} & 0.96         & 0.99         &    0.95    \\
AF      & 0.24         & 0.25         & 0.32          & 0.3    & 0.28          & 0.23         & 0.35      & 0.26         & 0.24         & 0.37         & 0.33   \\
BM      & 0.95         & 0.99         & 0.98          & 0.99   & 0.99          & \textbf{1.0} & 0.72      & 0.99         & \textbf{1.0} & 0.95         & 0.96   \\
CR      & \textbf{1.0} & \textbf{1.0} & 0.99          & 0.97   & \textbf{1.0}  & 0.99         & 0.96      & \textbf{1.0} & 0.98         & 0.99         & 0.88   \\
ER      & 0.93         & 0.98         & 0.84          & 0.89   & \textbf{0.99} & 0.98         & 0.91      & 0.98         & 0.82         & 0.93         & 0.84   \\
EP      & 0.96         & 0.99         & 0.99          & 0.96   & \textbf{1.0}  & 0.99         & 0.67      & 0.99         & \textbf{1.0} & \textbf{1.0} & 0.94   \\
EC      & 0.3          & 0.45         & \textbf{0.82} & 0.29   & 0.79          & 0.67         & 0.31      & 0.48         & 0.49         & 0.6          & 0.32    \\
HD      & 0.3          & 0.45         & 0.35          & 0.32   & 0.4           & 0.46         & 0.27      & 0.45         & 0.28         & 0.35         & 0.27   \\
HW      & 0.61         & 0.57         & 0.29          & 0.33   & 0.56          & 0.34         & 0.34      & 0.55         & 0.18         & 0.54         & 0.2    \\
LI      & 0.88         & 0.91         & 0.84          & 0.84   & 0.93          & 0.91         & 0.79      & 0.89         & 0.82         & 0.87         & 0.71   \\ \arrayrulecolor{black!30} \specialrule{0.07em}{0pt}{0pt}
Mean    & 0.72         & 0.76         & 0.74          & 0.69   & 0.79          & 0.76         & 0.63      & 0.76         & 0.68         & 0.76         &   0.64  \\ 
\arrayrulecolor{black} % \bottomrule
\end{tabular}
\qquad \qquad % Space between tables
\begin{tabular}{@{}lcccccccccc!{\color{black!30}\vrule width 0.07em}c>{\columncolor[HTML]{dcdcdc}[1pt][1pt]}c@{}} 
\toprule
Dataset & MUSE & InceptionT & HC1  & cBOSS & ResNet & TSF  & TDE  & DTW-A & CIF  & RSF  & Mean & \textbf{STELLE} \\  \specialrule{0.07em}{0pt}{0pt}
AW      & 0.99 & 0.99       & 0.98 & 0.98  & 0.98   & 0.95 & 0.98 & 0.99      & 0.98 & 0.98 & 0.98 &   0.95     \\
AF      & \textbf{0.74} & 0.22       & 0.29 & 0.3   & 0.36   & 0.3  & 0.3  & 0.22      & 0.25 & 0.28 & 0.31 & 0.33   \\
BM      &\textbf{1.0} &\textbf{1.0}       &\textbf{1.0} & 0.99  &\textbf{1.0}   &\textbf{1.0} & 0.99 &\textbf{1.0}      &\textbf{1.0} &\textbf{1.0} & 0.98 & 0.96   \\
CR     &\textbf{1.0} & 0.99       & 0.99 & 0.98  & 0.99   & 0.93 & 0.99 &\textbf{1.0}      & 0.98 & 0.97 & 0.98 & 0.88   \\
ER    & 0.97 & 0.92       & 0.94 & 0.84  & 0.87   & 0.9  & 0.94 & 0.93      & 0.96 & 0.92 & 0.92 & 0.84   \\
EP     &\textbf{1.0} & 0.99       &\textbf{1.0} &\textbf{1.0}  & 0.99   & 0.97 &\textbf{1.0} & 0.97      & 0.98 & 0.96 & 0.97 & 0.94   \\
EC      & 0.49 & 0.28       & 0.81 & 0.4   & 0.29   & 0.45 & 0.53 & 0.3       & 0.73 & 0.34 & 0.49 & 0.32    \\
HD     & 0.38 & 0.42       & 0.38 & 0.29  & 0.35   & 0.49 & 0.38 & 0.31      & \textbf{0.52} & 0.32 & 0.37 & 0.27  \\
HW     & 0.52 & \textbf{0.66}       & 0.5  & 0.49  & 0.6    & 0.36 & 0.56 & 0.61      & 0.35 & 0.37 & 0.47 & 0.2    \\
LI      & 0.9  & 0.89       & 0.9  & 0.85  & \textbf{0.94}   & 0.8  & 0.88 & 0.88      & 0.92 & 0.76 & 0.87 &   0.71     \\\arrayrulecolor{black!30} \specialrule{0.07em}{0pt}{0pt}
Mean   & \textbf{0.8 } & 0.74       & 0.78 & 0.71  & 0.74   & 0.72 & 0.76 & 0.72      & 0.77 & 0.69 & 0.73 &  0.64   \\ \arrayrulecolor{black} \bottomrule
\end{tabular}
\end{table}

We evaluate STELLE on ten benchmark datasets from the UEA Time Series Archive \citep{bagnall2018ueamultivariatetimeseries}, comparing it against both state-of-the-art and classical baselines introduced in earlier. The evaluation follows a protocol of 10 folds with 3 seeds each (30 total runs per dataset), consistent with the experimental setup used to obtain the baseline performances reported in \citet{Ruiz_Flynn_Large_Middlehurst_Bagnall_2021}.
All experiments were executed on a DELL PowerEdge R7525 server equipped with dual AMD EPYC 7542 CPUs (32 cores each), 768~GB RAM, and two NVIDIA A100 GPUs. The GPU resources were partitioned, with each run typically using up to 10~GB of GPU memory, and at most 20~GB for the largest datasets. CPU memory usage ranged between 20~GB and 150~GB, with only two datasets requiring allocations above 50~GB. % demetra
% All models were tuned for predictive accuracy only, as detailed in Appendix \ref{app:results}.
Across datasets, STELLE achieves moderate yet consistent performance. While it does not rank among the top-performing methods in terms of raw accuracy, it maintains stable results across diverse domains, including motion capture, physiological signals, and spectrographic data. For instance, on datasets such as \textit{ArticularyWordRecognition} (AW) and \textit{Epilepsy} (EP), its performance approaches that of ensemble-based methods such as HC2 and DrCIF, whereas on more complex, high-dimensional tasks, such as \textit{HandMovementDirection} (HD) and \textit{EthanolConcentration} (EC) its accuracy remains competitive despite the additional interpretability constraints.

It is worth emphasising that STELLE is the only inherently interpretable method among all compared approaches. Whereas existing neural and ensemble models (e.g., ROCKET, InceptionTime, ResNet, HC2) provide no direct insight into their decision processes and require external post-hoc explainers, STELLE produces symbolic explanations natively, in the same formal language, Signal Temporal Logic, used for reasoning. Moreover, to the best of our knowledge, it is the first model to provide both local (trajectory-specific) and global (class-level) logical explanations for multivariate time series.

Although a trade-off between interpretability and accuracy is expected, STELLE demonstrates that meaningful, formally grounded explanations can be obtained without a substantial loss in predictive performance if compared with black-box SOTA algorithms. Its stability across datasets suggests that the proposed robustness-based representation captures general temporal properties that extend beyond dataset-specific patterns.

Table \ref{tab:smallresults} provides a comprehensive comparison of classification accuracy across all benchmark datasets, with STELLE's performance highlighted for direct comparison against these state-of-the-art methods. The table shows mean accuracy rankings that demonstrate STELLE's positioning within the current TSC landscape.

\subsection{Explanatory performance}
In addition to evaluating predictive accuracy, we assessed the interpretability of the explanations generated by STELLE for three representative datasets, \textit{BasicMotions} (BM), \textit{ERing} (ER), and \textit{Epilepsy} (EP), chosen as examples where STELLE’s analysis completed most efficiently. We fixed the parameters introduced in Section~\ref{sec:explextraction} $\bar \gamma_t = 0.8$ to retain only the most relevant concepts and $\bar \gamma_\ell = \bar \gamma_g =0$ in order to prioritise classification accuracy over interpretability. This configuration maximises local separability (Equation~\ref{eq:sepl}), allowing us to analyse the structural properties of the formulae without interpretability constraints. In particular, all learned formulae achieve perfect local separability scores in all evaluated cases.

Tables~\ref{tab:read_loc_pre} and \ref{tab:read_loc_post} report, respectively, the mean number of syntactic nodes and distinct variables in the local explanations before and after postprocessing, computed across all runs used to extract the classification accuracy reported earlier, distinguishing between correctly classified trajectories (\textit{C}), misclassified trajectories with explanations extracted with respect to the predicted label (\textit{Pred IC}), and misclassified trajectories with explanations extracted with respect to the true label (\textit{True IC}). 
As shown in the tables, correctly classified trajectories yield concise and semantically coherent formulae.
For misclassified instances, they tend to become substantially more complex and lengthy. This suggests that the model resorts to more elaborate temporal conditions when its internal representation is less confident or misaligned with the data, requiring more intricate rules to justify its decisions.

The postprocessing step, which refines formula thresholds and simplifies redundant logical structures, consistently reduces the average formula length while maintaining semantic integrity and discriminative power. As expected, the number of variables remains stable across pre- and postprocessing, since simplification only affects formula composition rather than the underlying signal dimensions. These results confirm the effectiveness of our rescaling and simplification procedures in enhancing readability without compromising explanation fidelity. 

\begin{table}
\caption{Readability statistics of local explanations before postprocessing. ``C'' denotes correctly classified samples; ``\textit{Pred IC}'' and ``true IC'' refer to explanations constructed with respect to predicted and true labels of misclassified samples, respectively. ``nodes'' is the average number of syntactic operators per formula, and ``vars'' indicates the number of distinct variables used.}
\centering
\label{tab:read_loc_pre}
\begin{tabular}{@{}lcccccc@{}}
\toprule
Dataset      & C nodes & Pred IC nodes & True IC nodes & C vars & Pred IC vars & True IC vars \\ \midrule
BasicMotions (BM) & 9.55 ± 12.43  & 20.33 ± 13.20        & 14.67 ± 5.19         & 2.08 ± 1.35  & 4.00 ± 1.41         & 3.67 ± 0.94         \\ 
ERing (ER)       & 4.07 ± 2.20   & 5.23 ± 2.54          & 5.18 ± 3.44          & 1.30 ± 0.49  & 1.59 ± 0.59         & 1.51 ± 0.55         \\
Epilepsy (EP)    & 10.41 ± 10.89 & 12.00 ± 7.48         & 39.67 ± 35.12        & 1.58 ± 0.72  & 2.33 ± 0.94         & 2.00 ± 0.82         \\ \bottomrule
\end{tabular}
\end{table}

\begin{table}
\caption{Readability statistics of local explanations after postprocessing. Postprocessing consistently reduces syntactic length while preserving the relative complexity difference between correct and incorrect predictions.}
\centering
\label{tab:read_loc_post}

\begin{tabular}{@{}lcccccc@{}}
\toprule
Dataset      & C nodes & Pred IC nodes & True IC nodes & C vars & Pred IC vars & True IC vars \\ \midrule
BasicMotions (BM)& 9.07 ± 11.25  & 19.00 ± 11.31        & 14.33 ± 4.71         & 2.08 ± 1.35  & 4.00 ± 1.41         & 3.67 ± 0.94         \\
ERing (ER)       & 3.78 ± 2.16   & 4.74 ± 2.32          & 4.69 ± 3.01          & 1.30 ± 0.49  & 1.59 ± 0.59         & 1.51 ± 0.55         \\
Epilepsy (EP)    & 9.20 ± 9.71   & 9.67 ± 6.18          & 34.00 ± 31.95        & 1.58 ± 0.72  & 2.33 ± 0.94         & 2.00 ± 0.82         \\ \bottomrule
\end{tabular}
\end{table}

Global explanations, summarised in Table~\ref{tab:global}, follow similar patterns. Each table entry includes the global separability scores, as defined in Equation~\ref{eq:sep}, filtered by correctly (\textit{Sep C}) and incorrectly (\textit{Sep IC}) classified trajectories, as well as for all trajectories combined (\textit{Sep}), encompassing both correct and incorrect classifications. Moreover, we report the average number of syntactic nodes and variables of the global classifier before (\textit{Readability Pre}) and after (\textit{Readability Post}) postprocessing, summarising the overall compactness, indicated by the number of nodes, and dimensional coverage, indicated by number of variables, of the class-level formulae. % These results indicate that the selected logical formulae effectively capture discriminative temporal patterns while mostly maintaining a compact and interpretable structure across datasets. 
Lastly, the table presents \textit{Recall}, \textit{Specificity} and \textit{Precision} scores of the global explanation-based classifier. These metrics measure the extent to which the extracted logical rules successfully replicate the neural network's classification performance, providing insight into whether the explanations capture the essential decision boundaries learned by the neural network.

The global separability scores remain relatively high across datasets, confirming that the selected logical formulae effectively capture discriminative temporal patterns. \\
Interestingly, separability scores for incorrectly classified (IC) cases are most times higher than for correct ones, though these explanations tend to be substantially more complex, as reflected in their higher node counts. This suggests that the model continues to identify discriminative patterns in misclassified cases, but expresses them through more complex logical formulae, which are consequently more difficult to interpret. Readability analysis further shows that postprocessing effectively reduces formula length while maintaining a stable number of variables across all conditions. Finally, the global recall, specificity, and precision values indicate that class-level explanations are both selective and faithful to the model’s decision boundaries: high specificity confirms that global concepts rarely activate for unrelated classes, while moderate recall and precision reflect a balanced coverage of relevant temporal features without overfitting to class noise. 
Together, these results demonstrate that STELLE’s local and global explanations remain discriminative, interpretable, and consistent with model behaviour across domains. The combination of symbolic interpretability and practical efficiency reinforces its potential as a trustworthy neuro-symbolic framework for time series classification.

\begin{table}
\caption{Global explanation separability and readability. ``Sep'' corresponds to global separability (as defined in Equation~\ref{eq:sep}) computed on correctly classified (C), misclassified (IC), or all samples. ``Readability" reports the mean ± standard deviation of syntactic nodes (n) and variables (v). ``Spec'' is specificity, ``Prec'' is precision.}
\centering
\label{tab:global}
\setlength{\tabcolsep}{2pt} 
\begin{tabular}{@{}lcccccccc@{}}
\toprule
Dataset      & Sep C (\%) & Sep IC (\%) & Sep (\%)  & Readability Pre       & Readability Post      & Recall & Spec & Prec \\ \midrule
BasicMotions (BM) & 59.9\%           & 76.4\%             & 60.6\%       & 46.2±42.8n, 3.0±2.1v  & 30.2±26.8n, 3.0±2.1v  & 50.0\% & 97.5\%      & 87.0\%    \\
ERing (ER)       & 74.5\%           & 40.9\%             & 71.4\%       & 8.7±3.2n, 2.0±0.6v    & 8.7±3.2n, 2.0±0.6v    & 66.7\% & 98.7\%      & 98.9\%    \\
Epilepsy (EP)    & 72.2\%           & 95.9\%             & 71.8\%       & 106.0±53.8n, 3.0±0.0v & 101.8±52.0n, 3.0±0.0v & 74.6\% & 93.9\%      & 80.5\%    \\ \bottomrule
\end{tabular}
\end{table}

Computation times vary substantially across datasets but remain practical, with inference consistently completing in under one minute and full training with explanation extraction never exceeding one hour on the above-mentioned setup. Overall, these results demonstrate that STELLE produces concise, fully separable, and computationally tractable logical explanations, bridging symbolic reasoning with time series classification.

\begin{figure}
\centering
\begin{subfigure}{0.49\textwidth}
    \centering
    \includegraphics[width=\linewidth]{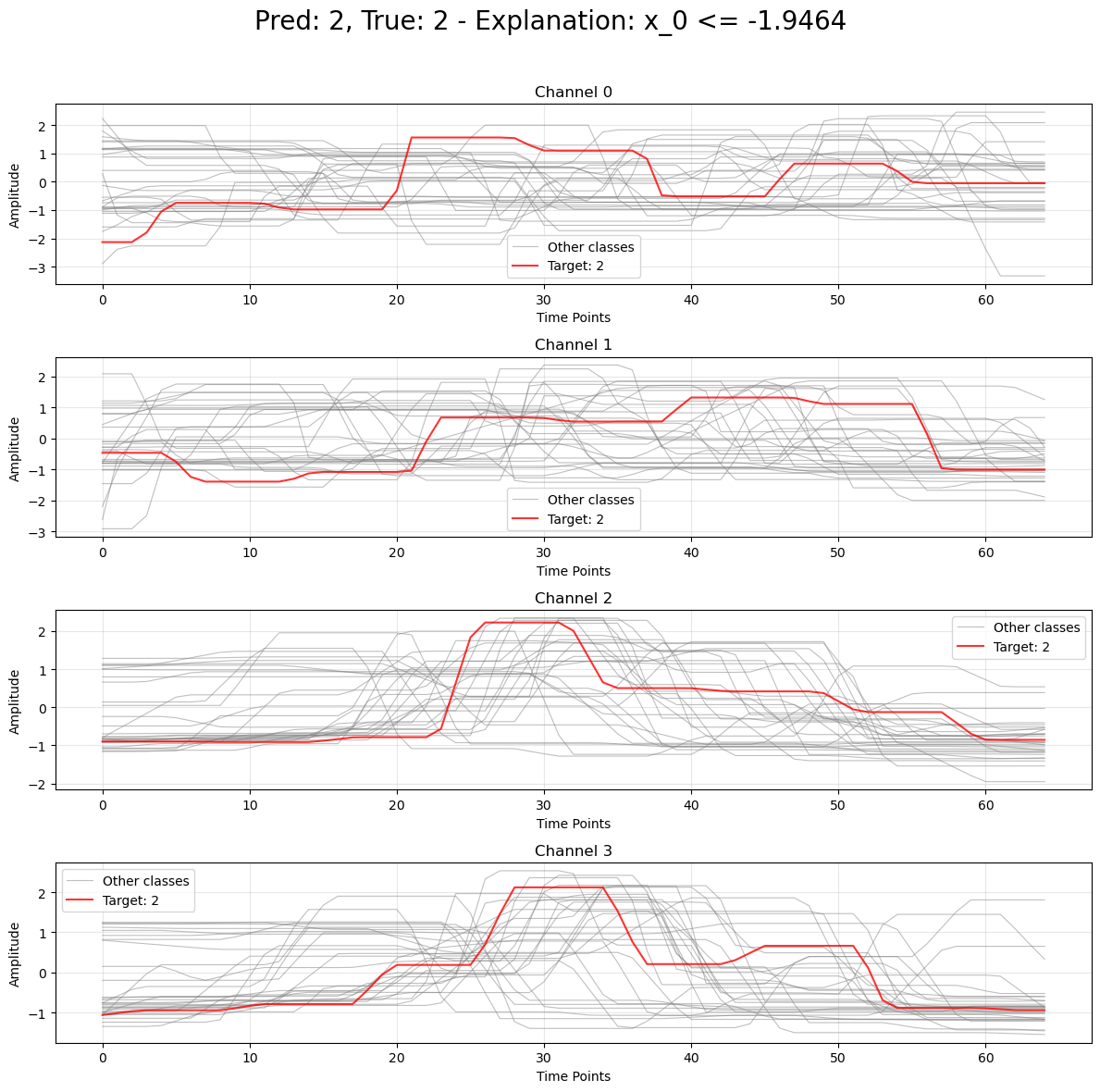}
\end{subfigure}
\hfill
\begin{subfigure}{0.49\textwidth}
    \centering
    \includegraphics[width=\linewidth]{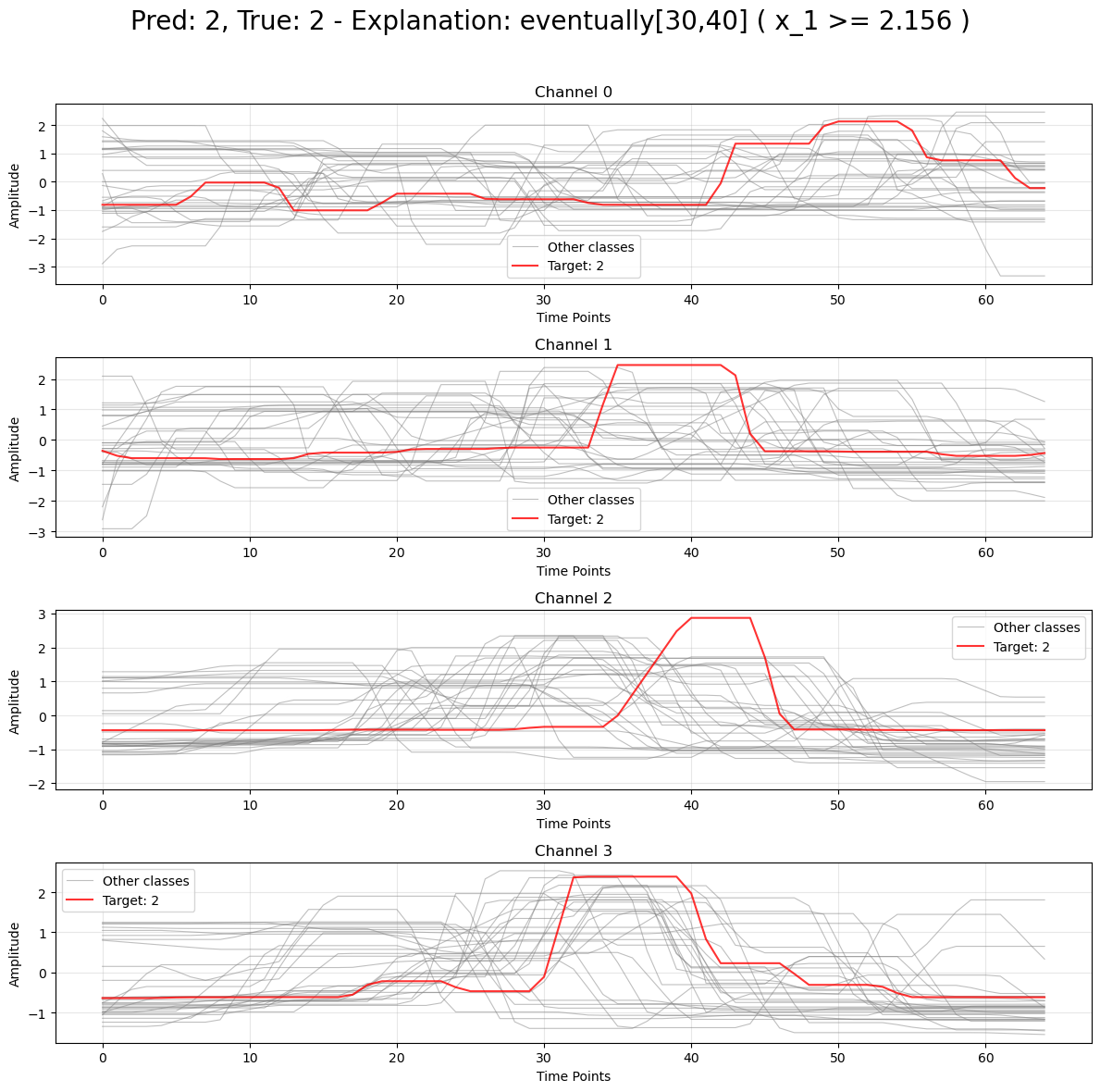}
\end{subfigure}
\caption{Local explanations for two trajectories in the \textit{ERing} dataset, showing the target trajectory (in red) and the trajectories from other classes (in gray) across its four channels. As title, the true and predicted class, and the proposed postprocessed explanation.}
\label{fig:locexample}
\end{figure}

\begin{figure}
    \centering
    \includegraphics[width=0.85\linewidth]{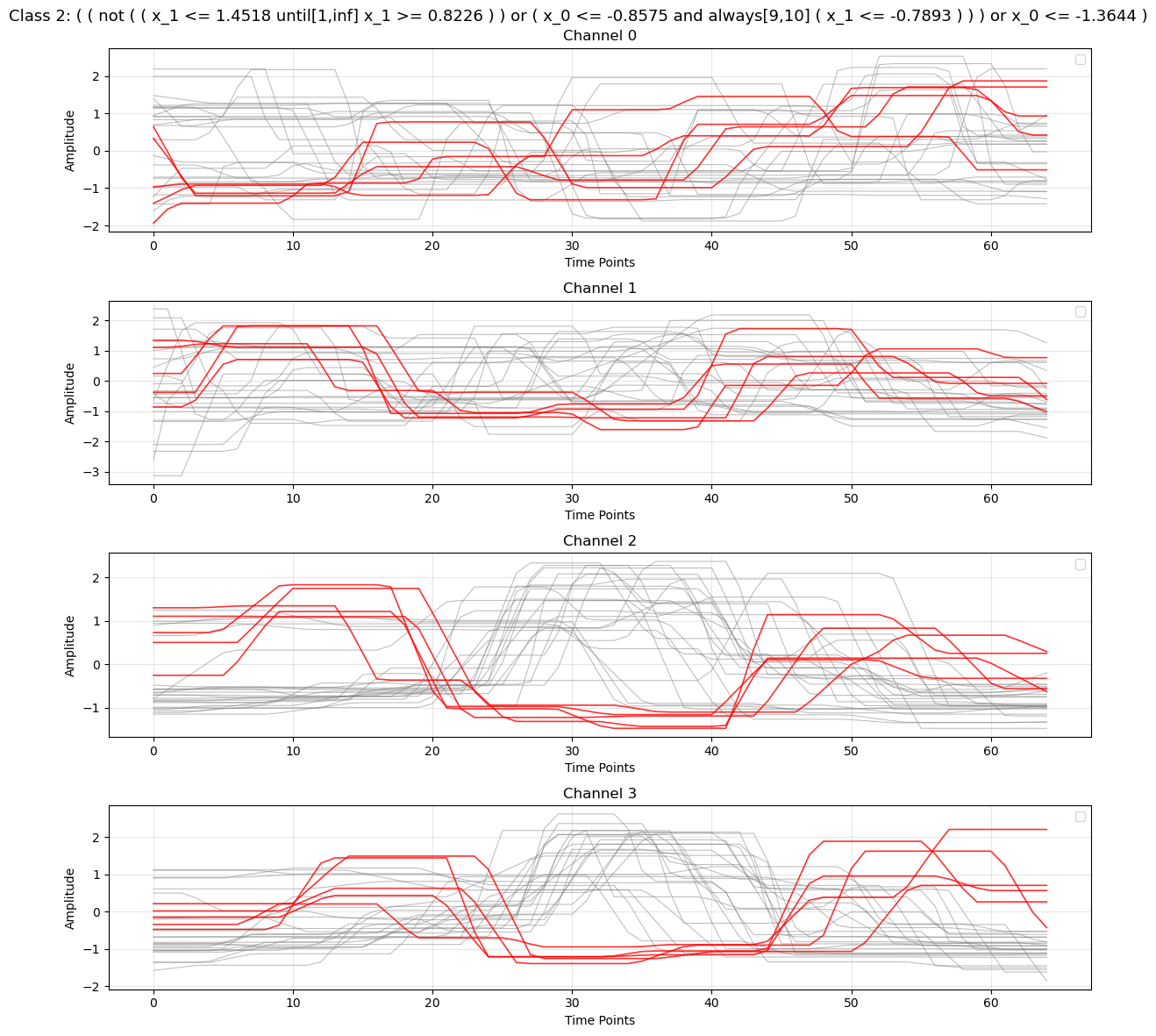}
    \caption{Global explanation for class 2 in the \textit{ERing} dataset, showing the target trajectories (in red) taken from the test set wrt their predicted class,  and the training trajectories from other classes (in gray) across its four channels. As title, the class and the proposed postprocessed explanation.}
\label{fig:globexample}
\end{figure}

As a practical example, we report in Figure~\ref{fig:locexample} and Figure~\ref{fig:globexample} a subset of local explanations and the global explanations, respectively, generated for the dataset \textit{ERing}, which contains $6$ classes and $4$ variables. 
\\The local explanations (Figure~\ref{fig:locexample}) showcase STELLE's ability to identify instance-specific temporal patterns that drive classification decisions. The explanations are expressed as STL formulae with temporal operators and threshold conditions on specific features:
\begin{align*}
&x_0 \leq -1.9464\\
&\mathbf{F}_{[30,40]} \ (x_1 \geq 2.156)
\end{align*}
This provides human-readable rules in the form of natural language-like statements that are accessible to all, regardless of their technical background or domain expertise.\\
The visualisations provide evidence of the explanations' validity: the red highlighted time series, the targets, clearly exhibit the patterns described in the explanations. In the first instance, feature $x_0$ remains below the threshold throughout the early time window, while in the second instance, feature $x_1$ spikes above $2.156$ in the time window $[30,40]$, exactly as specified by the \textit{Eventually} operator. 

The global explanation (Figure~\ref{fig:globexample}) provides a class-level understanding for class 2 that generalises across multiple instances of the same class shown in the local explanations. The formula reveals three distinct temporal pathways through which a time series can be classified as class 2, which can be interpreted as
\begin{align*}
&\mathbf{F}_{[12,19]} \ (x_1 \geq 1.8418) \\
&\mathbf{F}_{[12,19]} \ (x_1 \geq 1.8418)  \ \text{and} \  ( \mathbf{G}_{[29,38]} \ (x_2 \leq -0.33)\\
&\mathbf{G}_{[20,22]}( \mathbf{F}_{[3,33]} (x_3 \leq -0.0514) \ \text{and} \ x_3 \geq 1.5653))
\end{align*}
These conditions, connected by logical OR operators, indicate that a time series belongs to class 2 if it satisfies at least one of these three temporal patterns.\\
This disjunctive structure demonstrates STELLE's ability to capture the diverse temporal patterns that characterise a single class. The multiple trajectories visible in red show how different instances of class 2 can satisfy different parts of the global formula while still belonging to the same class. This flexibility is crucial for capturing the natural variability present in real-world time series data.

These results demonstrate STELLE's dual contribution to explainable time series classification. First, local explanations allow to verify individual predictions, enabling debugging and identification of potential errors or biases in specific instances. Second, global explanations reveal temporal patterns at the class level, potentially uncovering domain insights about the underlying processes generating the time series.

\section{Related work} \label{sec:rw}

\paragraph{Explainable AI (XAI) for TSC} TSC involves learning a classifier which associates trajectories with a probability distribution over discrete class labels. It is a traditional problem in ML \citep{tsc-traditional-survey,Ruiz_Flynn_Large_Middlehurst_Bagnall_2021} with DL approaches being the state of the art in this field \citep{dl-tsc}. Among them generative models are becoming popular for TSC, recently being enriched by foundation models for time series data \citep{ts-foundational}, built on top of the Transformer architecture \citep{vaswani2023attention}, to learn versatile representations of trajectories from vast amounts of data. XAI in the TSC context mostly adapts techniques from other domains to time series \citep{tsc-posthoc-survey,tsc-xai-survey}. Recent surveys \citep{rojat2021explainable, šimić2021xaimethodsneuraltime, Arsenault_2025} underscore the growing need for time-series-specific explainability, especially in high-risk domains. Toolkit developments such as Tsinterpret \citep{höllig2022tsinterpretunifiedframeworktime} and evaluations of perturbation robustness \citep{schlegel2023deepdiveperturbationsevaluation} highlight practical and reliability challenges in time series XAI. 
Attribution methods are arguably the most prominent techniques, aiming to identify time points or sub-sequences in the input that most influenced the prediction. Attention-based explanations, for instance, leverage attention mechanisms \citep{Karim_2018, vinayavekhin2018focusing, ge2018interpretable} to assign relevance scores to parts of the sequence. Extensions of this idea to multivariate settings include models that simultaneously learn to attend to both important time intervals and variables \citep{hsieh2020explainable}, or dual-mode architectures designed for interpretability across feature and temporal dimensions \citep{cai2024explainable}.
\\ Beyond temporal attribution, several works have proposed alternative interpretability strategies. Some focus on learning inherently interpretable representations of trajectories, such as the prototype-based latent space introduced in \citet{baldan2021interpretable}. Others explore post-hoc or counterfactual explanations: \citep{refoyo2024multi} proposes sparse, subsequence-based counterfactuals that highlight minimal changes needed to alter the prediction, while ensemble-based approaches have been used to identify robust, medically-relevant features driving predictions in multivariate time series \citep{metsch2025ensemble}. Functional decomposition \citep{degraff} offers a novel route to interpretable feature-level explanations by aligning latent behaviour with meaningful subsequence dynamics. Symbolic approaches such as EMeriTAte+DF \citep{bergami2025explainablesequentiallearning} show that hybrid numeric-event reasoning can improve both verifiability and explainability in multivariate settings. While EMeriTAte+DF generates explanations through hybrid numeric-event detection (e.g., identifying statistical anomalies as events), STELLE advances this paradigm by fully grounding explanations in temporal logic semantics, representing concepts as STL formulae that explicitly encode temporal relationships and enabling direct logical verification of model decisions.

% Among them, attribution methods are arguably the most famous and aim at finding the time points or sub-sequences inside the input which have influenced the prediction the most; among them, attention-based explanations leverages attention mechanisms \citep{Karim_2018, vinayavekhin2018focusing, ge2018interpretable} inside the classifier to assign a relevance score to different parts of the input sequence. 

\paragraph{Learning temporal logic formulae as TS classifiers} Learning temporal logic formulae as TS classifiers has recently received notable attention. Pioneer works \citep{bombara,roge,enumerate-stl} cast the TSC problem as a data-driven requirement mining task, and use the evaluation of the mined properties as classifier. 
Related symbolic approaches, such as those by \citet{kadous}, have historically explored parameterised event primitives to extract global, rule-based classifiers from time series data, anticipating more recent developments in symbolic temporal reasoning.
% leverage Parametric STL (PSTL), i.e. an extension of STL which replaces time bounds and constants of STL formulae with free parameters, and learns the parameters of such PSTL templates in a data-driven manner, using the resulting STL formula as classifier. 
% In particular in \citep{bombara} a decision tree classifier is learned, having as nodes some predefined parametric STL primitives, using ad-hoc impurity measures; classification is done by traversing the tree, i.e. by the formula resulting reading the path corresponding to the input.
More recently NeSy models have emerged in this context, which can be roughly divided in: (i) methods combining a decision tree approach for TSC, in which nodes of the tree represent STL formulae, where either activation functions mimic robust semantics, or time series features are extracted using Neural Networks (NN) to decide how to split the tree nodes \citep{nesy-tsc-one,nesy-tsc-two}; and (ii) works directly translating the syntax tree of a temporal formulae into a NN, with neurons corresponding to different operators, and activation functions specifically devised towards solving TSC \citep{nesy-tsc-threev2,nesy-tsc-four}.
%. in \citep{nesy-tsc-three} each layer of the NN represents different sets of STL operators, having differentiable robustness as activation, and the final STL classifier is devised by checking which of them are active; in \citep{nesy-tsc-four} a similar procedure is proposed, but using wSTL and designing special activation functions.
% in \citep{nesy-tsc-one} weighted STL (wSTL) formulae are organized in a decision tree, and their weights are learned by means of activation functions which resemble wSTL robust semantics; in \citep{nesy-tsc-two} instead neural computations are injected into decision tree, by using a Neural Network (NN) for extracting time series features to decide how to split the tree nodes. Finally, another set of works directly translate the syntax tree of a temporal formulae into a NN, with neurons corresponding to different operators: in \citep{nesy-tsc-three} each layer of the NN represents different sets of STL operators, having differentiable robustness as activation, and the final STL classifier is devised by checking which of them are active; in \citep{nesy-tsc-four} a similar procedure is proposed, but using wSTL and designing special activation functions. 
In all these works however, the learned temporal formulae represent global, i.e. class-wise, explanations, as opposed to STELLE which natively computes local explanations, from which global ones can be naturally inferred. 
%  Moreover, our model maps both STL formulae and the input trajectories in a common latent space, where cross attention is done to assign relevance to each STL concept: this improves flexibility in concept selection, hence diversity of the explanations, yet keeping them as concise as possible. 
Our model evaluates a fixed set of human-interpretable STL formulae directly on the input trajectories. This direct, concept-based embedding allows explanation to be decoupled from fixed prototypes or latent structures, promoting diversity in the explanations while keeping them concise and logically grounded.

\paragraph{Concept-based models for TS data} Concept-based models for TS data did not receive much attention, with some notable exceptions: concepts extracted by clustering raw time series and possibly combined with domain knowledge are used in \citet{ts-concept-one} for TSC and data analysis; prototypes, which are class-representative trajectories, have also been widely used in TSC to facilitate the identification of patterns and improve the interpretability of classification models \citep{ts-concept-two, ghosal2021multi, wang2025};
% prototypes, i.e. class-representative trajectories, are 
%extracted as centers of clusters computed in the latent space of an autoencoder architecture for TSC in \citep{ts-concept-two}; often used as well \citep{ts-concept-two, ghosal2021multi, wang2025};
the concept activation vector approach \citep{tcav} has instead been recast for TSC in \citet{ts-concept-three}, by using human-defined concepts. Some works also explore symbolic abstraction via interpretable rule sets or logical primitives (e.g., \citet{kadous}), but these typically rely on handcrafted features or predefined event templates, lacking the data-driven adaptability of concept-based neural models. % \citep{baldan2021interpretable} and \citep{ts-concept-two} explored latent prototype spaces, while our work introduces temporal logic concepts as interpretable anchors in a shared embedding space. 
Hence, to the best of our knowledge, ours is the first approach proposing a NeSy concept-based architecture for TSC, in which temporal logic concepts are automatically inferred without human intervention. 

\paragraph{Learning meaningful vector representations of TS}
is a longstanding goal in machine learning. Classical approaches, such as dynamic time warping (DTW) combined with dimensionality reduction techniques \citep{dtw}, and shapelet-based methods \citep{ye2009time, binarySTC} offer interpretable yet limited representations. More recent deep learning techniques like TS2Vec \citep{yue2022ts2vecuniversalrepresentationtime} and LETS-C \citep{kaur2024letscleveraginglanguageembedding} learn embeddings via contrastive or language-inspired mechanisms, focusing on predictive performance over interpretability. Recent work in self-explainable graph-based models \citep{garcia2025self} and verbalised spatio-temporal explanations \citep{amslaurea30767} suggests a shift towards integrating explainability into the architecture itself. 
%A comprehensive overview of modern approaches to time series classification is provided in \citep{foumani2023deeplearningtimeseries}. 
% In contrast, we propose a novel embedding of trajectories derived from their similarity to interpretable STL concepts, using a robustness-inspired kernel. This alignment with logical semantics enables both classification and explanation within a unified, interpretable framework.
\\

\section{Conclusions}
% foster knowledge discovery about the system 
We introduced a novel interpretable framework for time series classification that leverages concepts expressed in Signal Temporal Logic (STL) and embeds both input trajectories and logical formulae into a shared semantic space. By designing a robustness-based kernel for input trajectories, we enable a direct and interpretable mapping from raw time series to symbolic representations, ensuring consistency between the classification mechanism and its logical explanations.
\\Our neuro-symbolic architecture combines the expressive power of temporal logic with the flexibility of neural modelling, allowing predictions to be explained through human-understandable temporal relations. Explanations are extracted both locally, for individual predictions, and globally, as class-characterising logical formulae obtained through symbolic aggregation. This integration ensures that interpretability is intrinsic to the model design, rather than an external post hoc component.

Our experimental evaluation on a diverse set of multivariate time series benchmarks demonstrates that STELLE achieves competitive predictive performance compared to state-of-the-art black-box models, while uniquely providing symbolic explanations grounded in Signal Temporal Logic. Although it does not consistently rank among the top-performing methods in accuracy, STELLE maintains stable results across domains and offers a level of interpretability unattainable by existing neural or ensemble-based approaches.

% Through extensive experiments on both real-world and synthetic datasets, we demonstrate that our method not only achieves competitive classification accuracy but also provides faithful, concise, and semantically meaningful explanations. By grounding time series analysis in temporal logic, our work opens new directions for interpretable learning in safety-critical domains, where understanding the rationale behind a model’s prediction is as important as the prediction itself.
\paragraph{Future Work}
Future developments will focus on improving the framework's scalability and adaptability. We plan to reduce the computational overhead of concept generation through more efficient sampling and optimization techniques, enabling efficient reasoning with high-dimensional or long-horizon data. One promising direction is to extend STELLE with hierarchical or probabilistic temporal logics, which could enhance its expressivity while preserving interpretability. Furthermore, incorporating online and continual learning mechanisms would allow the provided explanations to evolve with non-stationary data, supporting long-term deployment in dynamic domains.

\paragraph{Societal implications}
STELLE’s interpretable logic-based explanations can enhance trust in safety-critical applications like medical diagnostics or autonomous systems, while reducing risks of black-box bias. However, simplified logical representations may overlook complex temporal patterns, requiring careful validation to prevent misinterpretation in high-stakes decisions.

\paragraph{Limitations}
While the proposed framework demonstrates strong performance and interpretability across diverse time series classification tasks, several limitations remain.
First, the generation and evaluation of large numbers of candidate STL formulae introduce a significant computational overhead during concept construction. This step scales with both the number of variables and the density of the parameter grid, which may limit applicability to very high-dimensional or long-horizon datasets.
Second, the model's interpretability depends on the comprehensibility of individual formulae. Although we constrain formula depth and variable count during the construction of atomic temporal patterns, the process of combining multiple formulae with different variables to generate local and global explanations can result in complex composite expressions where both the variable count and formula length increase substantially, potentially reaching a level of complexity that becomes difficult for humans to interpret and understand.

\begin{acks}
This study was carried out within the PNRR research activities of the consortium iNEST (Interconnected North-Est Innovation Ecosystem) funded by the European Union Next\-GenerationEU (Piano Nazionale di Ripresa e Resilienza (PNRR) \- Missione 4 Componente 2, Investimento 1.5 \- D.D. 1058 23\/06\/2022, ECS\_00000043). This manuscript reflects only the Authors’ views and opinions, neither the European Union nor the European Commission can be considered responsible for them. 
\end{acks}

%%
%% The next line prints the references.
\printbibliography

\appendix
\section*{Appendix}
\newtheorem*{trasl}{Preposition}

\section{Kernel evaluation}\label{app:kernel}

Two kernel functions that capture the similarity between STL formulae and between trajectories and formulae are defined respectively in Equation~\ref{eq:stl-kernel} and Equation~\ref{eq:traj-formula-kernel}.
Being the space $\mathcal{T}$ of trajectories infinite-dimensional, we should define a measure over it, for being able to evaluate the two equations via Monte Carlo approximation. For this reason we provide a sampling algorithm for such measure, denoted as $\mu_0$, detailed in Algorithm~\ref{alg:mu0}, which operates on piecewise linear functions over the interval $[a, b]$ (which is a dense subset of the set of continuous functions over $I\subseteq \mathbb{R}_{\geq 0}$).

\begin{algorithm}
\caption{$\mu_0$ for sampling a trajectory over the interval $[a, b]$}
\label{alg:mu0}
\begin{algorithmic}
\Require $\Delta$, $a$, $b$, $m'$, $m''$, $\sigma'$, $\sigma''$, $q$
\Ensure $\tau$ 
\State \Comment{sample the starting point} 
\State $\tau_0 \sim \mathcal{N}(m', \sigma')$
\State $\tau(t_0) \gets \tau_0$
\State \Comment{sample the total variation}
\State $K \sim (\mathcal{N}(m'', \sigma''))^2$
\State $y_1,\dots,y_{N-1} \sim \mathbb{U}([0, K])$
\State $y_0 \gets 0$, $y_n \gets K$
\State orderAndRename($y_0, \ldots, y_n$) 
\State \Comment{now $y_1 \leq y_2 \leq \dots \leq y_{N-1}$}
\State $s_0 \sim \text{Discr}(-1, 1)$
\For{$i = 0$ to $N-1$}
    \State $s \gets$ Binomial($q$) \Comment{$P(s = -1) = q$}
    \State $s_{i+1} \gets s_i \cdot s$
    \State $\tau(t_{i+1}) \gets \tau(t_i) + s_{i+1}(y_{i+1} - y_i)$
\EndFor
\end{algorithmic}
\end{algorithm}

Default parameters are set as $a=0$, $b=100$, $\Delta=1$, $m'=m''=0''$, $\sigma'=\sigma''=1$, $q=0.1$. Intuitively, $\mu_0$ makes \emph{simple} trajectories more probable, considering total variation and number of changes in monotonicity as indicators of complexity of signals. 
Note that although the feature space $\mathbb{R}^{\mathcal{T}}$ into which $\rho$ 
% (and thus Equation (\ref{eq:stl-kernel})) 
maps formulae is infinite-dimensional, in practice the kernel trick allows to circumvent this issue by mapping each formula to a vector of dimension equal to the number of formulae which are in the training set used to evaluate the kernel (Gram) matrix.

\section{Concept selection and embedding} \label{app:concept_selection}
To construct a flexible and interpretable vocabulary of temporal logic concepts, we generate a large collection of STL formulae through symbolic enumeration, exploiting Parametric Signal Temporal Logic (PSTL)~\cite{pstl}. PSTL is an extension of STL that parameterises both the time bounds in the temporal operators and the thresholds in inequality predicates. Given a PSTL template and a parameter configuration, an STL formula is induced by instantiating its free parameters with the provided values. \\
Formula templates are constructed by recursively applying logical and temporal operators up to a maximum of $M = 5$ syntactic nodes. In order to ensure human interpretability, we limit the number of signal dimensions used per formula to $N \in \{1, 2, 3\}$. We do not allow more dimensions since, in practice, STL formulae involving a large number of variables tend to be difficult to interpret and verify visually. 
%As discussed in Appendix~\ref{app:ablation}, we 
We found that restricting $N = 1$ already yielded strong performance across all metrics, and we retained this setting for the final experiments.

We explored two different strategies for generating formulae across multiple variables. Suppose a dataset has $4$ variables and $N = 1$ is fixed. One approach was to generate, for instance, $1000$ formulae by selecting the formulae per variable independently. An alternative approach involved generating $250$ generic formulae with a placeholder variable, then duplicating them across all variables in the dataset (e.g., $x > 3$ becomes $x_0 > 3$, $x_1 > 3$, $x_2 > 3$, $x_3 > 3$). Empirically, both strategies proved essentially equivalent in performance% (see Appendix~\ref{app:ablation}),
but we adopt the second method as our default, as it promotes structural consistency across dimensions and simplifies the generation pipeline.
%Rather than fixing the total number of concepts, we specify a target number of concepts \emph{per input variable}, which allows us to avoid biases arising from datasets with widely different dimensionalities. For example, without this strategy, a dataset with $500$ dimensions would yield only a handful of concepts per variable, while a univariate dataset could receive an unbalanced number of formulae. 
We empirically found that allocating $500$ concepts per variable offered a good trade-off between coverage and computational efficiency. A minimum threshold of $1000$ concepts is enforced to ensure sufficient semantic diversity, even for low-dimensional datasets.
\\
Each formula template $\varphi$ is instantiated over a dense grid of parameter configurations $\mathcal{P} = \{p_1, \ldots, p_{|\mathcal{P}|}\}$ to produce a large pool of fully specified formulae $\varphi(p)$. These are evaluated over a representative set of trajectories $\mathcal{T} = \{\tau_1, \ldots, \tau_{|\mathcal{T}|}\}$ using the quantitative STL semantics, yielding a robustness matrix $S \in \mathbb{R}^{|\mathcal{P}| \times |\mathcal{T}|}$, with entries
\begin{equation*}
S(i, j) = \rho(\varphi(p_i), \tau_j)
\end{equation*}
Each row of $S$ is a robustness signature that characterises the behaviour of a formula across trajectories.
\\
Concept selection proceeds incrementally: candidate formulae are sampled in batches and compared to the current concept set via cosine similarity between their robustness signatures. A formula is retained if its similarity to all previously selected concepts is below a fixed threshold $t \in (0, 1]$. The threshold $t$ directly controls the trade-off between behavioural diversity and conceptual granularity: lower values enforce stricter separation, resulting in more distinct concepts at the cost of longer selection time, while higher values yield denser coverage of the formula space with faster computation. We selected $t=0.99$ as a good trade-off between computational efficiency and conceptual diversity. % The effects of this parameter are further discussed in Appendix~\ref{app:ablation}. 

When multiple candidates exhibit high similarity to existing concepts, we apply a preference for syntactic simplicity, retaining the structurally minimal representative among them. This procedure continues until the desired number of formulae is selected. The final concept set provides a compact, semantically diverse, and interpretable basis for embedding, reasoning, and symbolic explanation.

% Since we aim to maintain flexibility and diversity in the logical concepts used both within the model and in the explanations it produces, we precompute a large collection of STL formulae from which the model can draw the most semantically informative and discriminative ones. To this end, we first enumerate all STL templates (i.e., parametrised STL formulae) constrained to have at most $M = 5$ nodes and to refer to no more than two out of $N$ variables (i.e., input signal dimensions). Allowing up to two variables per formula enables us to capture interactions across dimensions, while still preserving human interpretability. The details of the process are reported in Algorithm~\ref{alg:stl-templates}.

% Since we want to keep our model flexible in terms of the diversity of concepts used in its internals and provided as explanation, we proceed by pre-computing a vast amount of STL formulae, and successively let the model extract those considered more useful for the classification task. 
% Hence, we construct the STL formulae used as candidate concepts by fixing the maximum number $M$ of nodes and the maximum number of variables $N$ (i.e. the dimensionality of the input signals) allowed, then we enumerate all templates (i.e., STL formulae where constants are replaced by parameters) satisfying those constraints as detailed in Algorithm~\ref{alg:stl-templates}.

\begin{algorithm}
\begin{adjustbox}{minipage=\linewidth,scale=0.95}
\caption{Algorithm for generating STL formulae templates}
\label{alg:stl-templates}
\begin{algorithmic}
\Require $M$, $N$
\Ensure all\_phis 
% \Comment{\text{STL templates of formulae with max $N$ vars and $M$ nodes}}
\State $\text{all\_phis}\gets [ \, ]$
\State all\_phis.append(\texttt{generateAtomicPropositions}()) % \Comment{$x_i\leq 0$ or $x_i\geq 0$, $\forall i\leq N$}
\For{$2\leq m\leq M$}
\State \Comment{retrieve templates with $m-1$ nodes}
\State $\text{prev\_phis}\gets$\texttt{getPhisGivenNodes}($m-1$) 
\State \Comment{$F$, $G$, $\neg$}
\State $\text{unary\_ops}\gets$\texttt{expandbyUnaryOperators}(prev\_phis) 
\State all\_phis.append(unary\_ops)
\State \Comment{all pairs $(l, r): l+r=m, l\leq r$}
\State l\_list, r\_list $\gets$\texttt{getPairsGivenSum}($m$) 
\For{$(l, r)\in$ [l\_list, r\_list]}
\State l\_phis$\gets$\texttt{getPhisGivenNodes}($l$)
\State r\_phis$\gets$\texttt{getPhisGivenNodes}($r$)
\State \Comment{$\wedge$, $\vee$, $U$}
\State binary\_ops$\gets$\texttt{expandbyBinaryOpeators}(l\_phis, r\_phis) 
\State all\_phis.append(binary\_ops)
\EndFor
\EndFor
\end{algorithmic}
\end{adjustbox}
\end{algorithm}

\section{Formulae manipulation} \label{app:phisimple}
In this section, we describe the postprocessing techniques applied to STL formulae in order to improve their interpretability, discriminative power, and syntactic simplicity. These procedures operate either by modifying formula thresholds based on observed robustness values, or by rewriting the logical structure using standard equivalences. Together, they ensure that the extracted explanations are both semantically meaningful and syntactically concise.

\subsection{Threshold shift}
To ensure that selected STL formulae are interpretable and exhibit class--discriminative behaviour, we apply a postprocessing step based on robustness analysis. For each input trajectory and its associated STL formula, we evaluate the formula’s robustness over trajectories from all classes. If the robustness of the input trajectory lies outside the range of robustness values for the other classes, i.e., the formula separates the target class, we rescale the formula’s threshold by a factor derived from the target robustness and its closest counterpart from other classes, aiming to enhance the separation margin. 

If, instead, the formula does not separate the target class, we adapt its decision boundary by analysing its robustness landscape across all classes. Specifically, for each formula, we compare its robustness on the target trajectory with the robustness values computed over trajectories of other classes. When the robustness distribution indicates that the formula fails to distinguish the target class, we shift its threshold toward the midpoint between the target robustness and the closest opposing robustness value, thereby maximising inter--class separation while preserving the original temporal structure. In both cases, to maintain semantic consistency, if the resulting robustness for the target trajectory becomes negative, the formula is logically negated. The final postprocessed formulae are then simplified to ensure readability and syntactic minimality.
\\This process is described in Algorithm~\ref{alg:postprocess}. 

We now formally justify this behaviour by proving the following proposition, which establishes the linearity of robustness with respect to uniform threshold shifts.
\begin{trasl}
By uniformly modifying the thresholds of the variables in an STL formula $\varphi$ by a constant $\delta$, this adjustment will shift the robustness values linearly by $\delta$.
\end{trasl}
\begin{proof}
We now provide proof of this statement for all STL. By proving that the Preposition holds true for every Atom and STL operator, the proof comes naturally for more complex structures. \\

\emph{Step 1: Atomic Predicates} \\
Consider the atomic predicate $\pi(\mathbf{x}) = \big(f_\pi(\mathbf{x}) \geq 0\big)$ and a trajectory $\tau$. \\Its robustness value is:
$$\rho(\pi,\tau,t) = f_\pi(\tau(t))$$
If we adjust the threshold by a constant $\delta$, this corresponds to modifying the predicate to 
$$\pi(\mathbf{x}) = \big(f_\pi(\mathbf{x}) + \delta \geq 0\big) = \big(f_\pi({x}) \geq -\delta\big)$$ The new robustness value becomes:
$$\rho'(\pi,\tau,t) = f_\pi(\tau(t)) - \delta = \rho(\pi,\tau,t) - \delta$$

Thus, a uniform adjustment by $\delta$ decreases the robustness linearly by $\delta$. \\

\emph{Step 2: Temporal operators} \\
Next, let's consider the impact on more complex formulae built from atomic predicates. \\

Let's start from \emph{Negation}, whose definition is:
   $$\rho(\lnot\varphi,\tau,t) = -\rho(\varphi,\tau,t)$$
   The a uniform adjustment by $\delta$ results in:
   \begin{align*}
       \rho'(\lnot\varphi,\tau,t) &= -\left(\rho(\varphi,\tau,t) - \delta\right) \\
       & = -\rho(\varphi,\tau,t) + \delta \\
       &= \rho(\lnot\varphi,\tau,t) + \delta
   \end{align*}
   This shows that the robustness shifts linearly by $\delta$.\\
   
For \emph{Conjunction}, defined as
   $$\rho(\varphi_1 \land \varphi_2, \tau, t) = \min\left(\rho(\varphi_1, \tau, t), \rho(\varphi_2, \tau, t)\right)$$
   after uniform adjustment becomes:
   $$\rho'(\varphi_1 \land \varphi_2, \tau, t) = \min\left(\rho(\varphi_1, \tau, t) - \delta, \rho(\varphi_2, \tau, t) - \delta\right)$$
   Since $\min(a - \delta, b - \delta) = \min(a, b) - \delta$, we have:
   $$\rho'(\varphi_1 \land \varphi_2, \tau, t) = \rho(\varphi_1 \land \varphi_2, \tau, t) - \delta$$
   Again, the robustness shifts linearly by $\delta$.\\
   
To prove the property for the \emph{Until} operator, we again start from its definition
\begin{align*}
 &\rho(\varphi_1 \mathbf{U}_{[a, b]} \varphi_2, \tau, t) = \\ &\max_{{t' \in [t+a,t+b]}}\left(\min\left(\rho(\varphi_2, \tau, t'), \min_{{t'' \in [t,t']}} \rho(\varphi_1, \tau, t'')\right)\right)
\end{align*}
After uniform adjustment:
   \begin{align*}
   &\rho'(\varphi_1 \mathbf{U}_{[a, b]} \varphi_2, \tau, t) = \\ &\max_{{t' \in [t+a,t+b]}}\left(\min\left(\rho(\varphi_2, \tau, t') - \delta, \min_{{t'' \in [t,t']}} \rho(\varphi_1, \tau, t'') - \delta\right)\right)
\end{align*}
   Simplifying using properties of $\max$ and $\min$:
   $$\rho'(\varphi_1 \mathbf{U}_{[a, b]} \varphi_2, \tau, t) = \rho(\varphi_1 \mathbf{U}_{[a, b]} \varphi_2, \tau, t) - \delta$$
   Hence, the robustness again shifts linearly by $\delta$. \\

Finally, the proof for \emph{Eventually} and \emph{Globally} comes naturally from their definitions and the properties of $\max$ and $\min$. \\

This concludes the proof that for any formula $\varphi$, the uniform adjustment of thresholds by a constant $\delta$ consistently shifts linearly the robustness value $\rho(\varphi, \tau, t)$. 
\end{proof}

\subsection{Pruning}
A summary of the simplifications applied can be found in Table~\ref{tab:simplifications}.
\subsubsection{Logical simplifications}
This simplification procedure applies structural equivalences to STL formulae. Double negation is eliminated via $\lnot(\lnot\varphi) \rightarrow \varphi$, while De Morgan laws transform $\lnot(\varphi \land \psi)$ into $\lnot\varphi \lor \lnot\psi$ and $\lnot(\varphi \lor \psi)$ into $\lnot\varphi \land \lnot\psi$. Redundant binary expressions are collapsed using $\varphi \land \varphi \rightarrow \varphi$ and $\varphi \lor \varphi \rightarrow \varphi$. The simplifier also reduces combinations like $\varphi \land (\varphi \lor \psi)$ and $\varphi \lor (\varphi \land \psi)$ to $\varphi$.

Temporal nesting is flattened. Considering $I$ and $J$ as two generic temporal intervals, $\mathbf{G}_I(\mathbf{G}_J(\varphi))$ becomes $\mathbf{G}_{I+J}(\varphi)$, and similarly $\mathbf{F}_I(\mathbf{F}_J(\varphi)) \rightarrow \mathbf{F}_{I+J}(\varphi)$. When both children of an until operator are equal, the formula $\varphi\ \mathbf{U}_I\ \varphi$ simplifies to $\varphi$. This procedure does not depend on data and only rewrites formulae based on their syntactic structure. 
% and is represented in Algorithm~\ref{alg:logical_simplify}.
\begin{table*}[h]
\centering
\caption{STL formulae simplifications}
\label{tab:simplifications}
\renewcommand{\arraystretch}{1.2}

\begin{minipage}[t]{0.40\textwidth}
\centering
\textbf{Logical simplifications}

\begin{tabular}{ll}
\toprule
\textit{Original} & \textit{Simplified} \\
\midrule
$\neg(\neg\varphi)$ & $\varphi$ \\
$\neg(x \leq c)$ & $x > c$ \\
$\neg(\mathbf{F}_I(\varphi))$ & $\mathbf{G}_I(\neg\varphi)$ \\
$\neg(\mathbf{G}_I(\varphi))$ & $\mathbf{F}_I(\neg\varphi)$ \\
$\neg(\varphi_1 \vee \varphi_2)$ & $\neg\varphi_1 \wedge \neg\varphi_2$ \\
$\neg(\varphi_1 \wedge \varphi_2)$ & $\neg\varphi_1 \vee \neg\varphi_2$ \\
$\varphi \wedge \varphi$ & $\varphi$ \\
$\varphi \vee \varphi$ & $\varphi$ \\
$\varphi \wedge (\varphi \vee \psi)$ & $\varphi$ \\
$\varphi \vee (\varphi \wedge \psi)$ & $\varphi$ \\
$\varphi \wedge (\varphi \wedge \psi)$ & $\varphi \wedge \psi$ \\
$\varphi \vee (\varphi \vee \psi)$ & $\varphi \vee \psi$ \\
$\mathbf{G}_I(\mathbf{G}_J(\varphi))$ & $\mathbf{G}_{I+J}(\varphi)$ \\
$\mathbf{F}_I(\mathbf{F}_J(\varphi))$ & $\mathbf{F}_{I+J}(\varphi)$ \\
$\varphi \ \mathbf{U}_I \ \varphi$ & $\varphi$ \\
\bottomrule
\end{tabular}
\end{minipage}
\hspace{1em}
\begin{minipage}[t]{0.40\textwidth}
\centering
\textbf{Data-aware simplifications}

\begin{tabular}{ll}
\toprule
\textit{Original} & \textit{Simplified} \\
\midrule
% $\varphi \rightarrow \texttt{True}$ if $\varphi$ is always satisfied \\
% $\varphi \rightarrow \texttt{False}$ if $\varphi$ is always violated \\
$\neg \texttt{True}$ & $\texttt{False}$ \\
$\neg \texttt{False}$ & $\texttt{True}$ \\
$\texttt{True} \wedge \varphi$ & $\varphi$ \\
$\texttt{False} \wedge \varphi$ & $\texttt{False}$ \\
$\texttt{True} \vee \varphi$ & $\texttt{True}$ \\
$\texttt{False} \vee \varphi$ & $\varphi$ \\
$\mathbf{G}_{[0,\infty)}(\texttt{True})$ & $\texttt{True}$ \\
$\mathbf{G}_{[0,\infty)}(\texttt{False})$ & $\texttt{False}$ \\
$\mathbf{F}_{[0,\infty)}(\texttt{True})$ & $\texttt{True}$ \\
$\mathbf{F}_{[0,\infty)}(\texttt{False})$ & $\texttt{False}$ \\
$\varphi \ \mathbf{U}_{[0, b]} \ \texttt{True}$ & $\varphi$ \\
$\varphi \ \mathbf{U}_{[a,b]} \ \texttt{True}$ , $a \neq 0 $& $\mathbf{G}_{[0,a]} \varphi$ \\
$\varphi \ \mathbf{U}_I \ \texttt{False}$ & $\texttt{False}$ \\
$\texttt{True} \ \mathbf{U}_I \ \varphi$ & $\mathbf{F}_I(\varphi)$ \\
$\texttt{False} \ \mathbf{U}_I \ \varphi$ & $\texttt{False}$ \\
\bottomrule
\end{tabular}
\end{minipage}

\end{table*}

\subsubsection{Data-aware simplifications}
This method uses the evaluation of atomic predicates over a trajectory set $\mathcal{T}$ to simplify STL formulae contextually. Each atom $\pi(\mathbf{x})$ is evaluated across all considered trajectories and at all time steps. If the predicate holds (resp. fails) for all samples at a given $t$, it is marked as true (resp. false). Otherwise, it is marked as undefined at that time step.

Given this truth map, each subformula is recursively reduced. An atom that is uniformly true (resp. false) is replaced by $\top$ (resp. $\bot$). Boolean simplification is then applied, collapsing expressions like $\top \land \varphi$ to $\varphi$ and $\bot \lor \varphi$ to $\varphi$.

\begin{figure}
\centering
\begin{minipage}{\columnwidth}
\begin{algorithm}[H]
\caption{Postprocess formulae based on robustness separation}
\label{alg:postprocess}
\begin{algorithmic}
\Require Set of formulae $\Phi$, target class $\hat{y}$, input trajectory $\mathbf{x}$, trajectories by class $\{\mathcal{T}_1, \dots, \mathcal{T}_{K-1}\}$
\Ensure Postprocessed formulae $\Phi'$
\State $\Phi' \gets \emptyset$
\For{$\varphi \in \Phi$}
    \State $r_{\text{target}} \gets \rho(\varphi, \mathbf{x})$
    \For{$k = 0$ to $K-1$}
        \If{$c \ne \hat{y}$}
            \State $\mathcal{R}_k \gets [\rho(\varphi, \tau) \text{ for } \tau \in \mathcal{T}_c]$
        \Else
            \State $\mathcal{R}_k \gets [r_{\text{target}}]$
        \EndIf
    \EndFor
    \State $\mathcal{R}_{\text{opp}} \gets \bigcup_{k \ne \hat{y}} \mathcal{R}_k$
    \State $n_{<} \gets \#\{r \in \mathcal{R}_{\text{opp}} \mid r < r_{\text{target}}\}$
    \State $n_{>} \gets \#\{r \in \mathcal{R}_{\text{opp}} \mid r > r_{\text{target}}\}$
    \If{$n_{>} > n_{<}$}
        \State $\varphi \gets \neg \varphi$
        \State $r_{\text{target}} \gets -r_{\text{target}}$
        \State $\mathcal{R}_{\text{opp}} \gets [-r \mid r \in \mathcal{R}_{\text{opp}}]$
    \EndIf
    \State $r_{\text{closest}} \gets \max\{r \in \mathcal{R}_{\text{opp}} \mid r < r_{\text{target}}\}$ % \Comment{or $r_{\text{target}}$ if none}
    \State $s \gets (r_{\text{target}} + r_{\text{closest}})/2$
    \State $\varphi' \gets \texttt{RescaleThresholds}(\varphi, -s)$
    \If{$\rho(\varphi', \mathbf{x}) < 0$}
        \State $\varphi' \gets \neg \varphi'$
    \EndIf
    \State append {$\varphi'$ to $\Phi'$}
\EndFor
\State \Return $\Phi'$
\end{algorithmic}
\end{algorithm}
\end{minipage}
\end{figure}

\section{Experimental setting} \label{app:training}

In this section, we provide a detailed account of the training procedures used to develop the model presented in this paper. This includes information on the datasets, data preprocessing steps, and hyperparameter settings. These details are intended to ensure the reproducibility of our results and to offer insights into the training processes that led to the model performances reported in the main text.
% In addition, we conducted a test to evaluate the scalability of our model with reference to the number of variables in a synthetic dataset. % Its results are reported in Subsection~\ref{app:scal}.

\subsection{Datasets and preprocessing} \label{sec:ds}
We base our experiments on a subset of time series classification tasks from the University of East Anglia (UEA) Time Series Archive \cite{bagnall2018ueamultivariatetimeseries}. This benchmark suite comprises over $140$ univariate and multivariate datasets spanning diverse domains, including sensor readings, motion capture, physiological signals, and spectrographic measurements. For computational feasibility, we consider the first ten datasets in alphabetical order from the archive, excluding \textit{DuckDuckGeese} due to hardware constraints. All datasets are loaded and standardised (zero mean and unit variance per channel), and processed using the \texttt{aeon} Python library \cite{aeon24jmlr}, which ensures consistency with other state-of-the-art time series classification benchmarks \cite{Bagnall_2020, Ruiz_Flynn_Large_Middlehurst_Bagnall_2021}.

As the datasets are multivariate, we perform a preprocessing step to remove uninformative or redundant input variables. In particular, we discard variables whose temporal traces are strongly correlated across the dataset. This is achieved by computing the Pearson correlation matrix over all variable pairs after flattening the time dimension. Pairs exceeding a correlation threshold of $0.9$ are reduced by keeping only the more informative variable (based on average absolute magnitude). This dimensionality reduction step improves training efficiency and robustness without compromising semantic diversity in the inputs.

We adopt the standard train/test splits provided by the \texttt{aeon} archive, which are consistent with previous evaluation protocols in the literature. However, for hyperparameter tuning, an internal development split was used to better estimate generalisation performance. 

\paragraph{Computing infrastructure}
All experiments were executed on a DELL PowerEdge R7525 server equipped with dual AMD EPYC 7542 CPUs (32 cores each), 768 GB RAM, and two NVIDIA A100 GPUs. The GPU resources were partitioned, with each run typically using up to 10 GB of GPU memory, and at most 20 GB for the largest datasets. CPU memory usage ranged between 20 GB and 150 GB, with only two datasets requiring allocations above 50 GB. All computations were performed using mixed CPU–GPU execution within a controlled HPC environment to ensure reproducibility and efficient resource utilisation.
\\Key dependencies included \texttt{CUDA} (12.8.0), \texttt{PyTorch} (2.2.1) \cite{pytorch}, and GPU-accelerated libraries for data processing and machine learning. The software stack also incorporated \texttt{scikit-learn} (1.6.1) \cite{scikit-learn} and \texttt{NumPy} (1.26.2) \cite{numpy} for numerical and statistical operations. \texttt{PuLP} (3.1.1) \cite{Mitchell2011PuLPA} was used to solve the ILP for global explanation extraction. As mentioned, \texttt{aeon} (1.1.0) \cite{aeon24jmlr} was used to handle the datasets.

\subsection{Training details}
All experiments were conducted using PyTorch with deterministic seed initialisation to ensure reproducibility. Each model was trained using the Adam optimiser with batch-wise updates and default momentum parameters. Weighted cross-entropy loss was used to account for class imbalance, where class weights were computed as the inverse frequency of each class within the training data. In addition to the standard loss, the final objective included regularisation terms to control the sharpness of the concept relevance distribution and the magnitude of the learned robustness threshold $\varepsilon$, as will be detailed later. Gradients were clipped within reasonable bounds, and concept relevance scale parameters were constrained during training to avoid numerical instability.

To better balance learning dynamics between the main model parameters and regularisation controllers, we define two parameter groups with distinct learning rates. 
The primary model parameters (e.g., weights, classification head, embeddings) are updated with a standard learning rate, while the specialised parameters governing regularisation dynamics, such as concept relevance collapse strength and $\varepsilon$ stability, are assigned a learning rate an order of magnitude higher, ensuring faster adaptation of these auxiliary components without destabilising the main optimisation process. As mentioned, both groups are optimised jointly using the Adam optimiser, allowing for robust and adaptive gradient updates across all layers.

Early stopping was employed with a configurable patience criterion during tuning, monitoring validation loss every few epochs. If no improvement was observed for a given number of consecutive validation checks, training was halted and the best model checkpoint (based on validation loss) was restored. % Robustness statistics (i.e., class-specific mean and standard deviation of each concept’s robustness) were precomputed before training, using only training data.

For each dataset, training was conducted on the official training split as provided by the \texttt{aeon} implementation of the UEA archive. 

\subsubsection{Hyperparameter tuning}
We employed a two–stage Bayesian optimisation strategy using the Ax framework \cite{ax} to efficiently explore the hyperparameter space. The search was performed on a dedicated fold distinct from those later used for official evaluation, following an $80/20/20$ train/validation/test split. Initial exploration was carried out using Sobol sampling to ensure broad coverage of the search space, followed by Bayesian optimisation leveraging qNoisyExpectedImprovement for single–objective tuning (accuracy) and qLogNParEGO or qNoisyExpectedHypervolumeImprovement \citep{nehvi} for multi–objective tuning (accuracy and local separability).

The search space spanned the number of layers ($0$–$3$), learning rate ($10^{-5}$–$10^{-1}$), initial concept relevance scale ($0.5$–$2.0$), and hidden layer dimension ($256$, $512$, $1024$). Other hyperparameters were fixed based on preliminary analysis: batch size ($32$), activation function (GELU), dropout rate ($0.1$), explanation threshold $\gamma_t = 0.8$, and concept generation parameters ($t = 0.99$, $500$ concepts per variable, with a minimum of $1000$ total concepts).

The optimisation consisted of an initial phase of $8$ Sobol trials followed by an extensive Bayesian refinement phase with early stopping based on validation performance. For the multi–objective setup, Pareto–optimal configurations were identified to balance predictive accuracy and local explanation quality. The final configuration was chosen according to test set performance in the tuning phase and subsequently evaluated on the official $10$–fold splits, training three independent seeds per fold. Reported results correspond to averages across all folds and seeds.

\subsubsection{Loss function} \label{app:loss}
The overall loss combines standard classification loss with regularisation terms that promote concept relevance interpretability and well-behaved robustness normalisation. It is defined as:

\begin{equation*}
\mathcal{L} = 
\underbrace{\text{CE}(\mathbf{z}, \mathbf{y}; \mathbf{w})}_{\text{classification loss}} 
+ 
\underbrace{\lambda_{T} \cdot \sigma\left(-\frac{T}{t}\right)}_{\text{concept relevance sharpness penalty}} 
+ 
\underbrace{\lambda_{\varepsilon} \cdot \left(e^{\varepsilon} + e^{-\varepsilon}\right)}_{\text{trajectory robustness regularisation}}
\end{equation*}

where:
\begin{itemize}
    \item $\text{CE}(\mathbf{z}, \mathbf{y}; \mathbf{w})$ is the weighted cross-entropy loss between the predicted logits $\mathbf{z} \in \mathbb{R}^K$ and one-hot encoded labels $ \mathbf{y} \in \{0, 1\}^K$, with class weights $ \mathbf{w} $,
    \item $  T$ is the scaling parameter used in Equation~\ref{eq:T},
    \item $ t $ is the concept relevance temperature,
    \item $ \varepsilon $ is the learnable trajectory robustness parameter (used in Equation~\ref{eq:epsilon}),
    \item $ \sigma(\cdot) $ is the sigmoid function,
    \item $ \lambda_{T}$ and $ \lambda_{\varepsilon} $ are trainable strengths for each regulariser.
\end{itemize}

The concept relevance sharpness term encourages concentrated distributions while avoiding instability through smooth sigmoid control. The robustness regulariser penalises extreme values of $\varepsilon$, promoting stable discriminability across concepts. Both regularisation terms are softly enforced via learnable scaling coefficients, allowing the model to dynamically balance interpretability and predictive performance during training.

\newpage
\section*{Reproducibility Checklist for JAIR}

Select the answers that apply to your research -- one per item. 

\subsection*{All articles:}

%\hh{revised for stylistic consistency:}
\begin{enumerate}
    \item All claims investigated in this work are clearly stated. 
    [yes/partially/no] Yes
    \item Clear explanations are given how the work reported substantiates the claims. 
    [yes/partially/no] Yes
    \item Limitations or technical assumptions are stated clearly and explicitly. 
    [yes/partially/no] Yes
    \item Conceptual outlines and/or pseudo-code descriptions of the AI methods introduced in this work are provided, and important implementation details are discussed. 
    [yes/partially/no/NA] Yes
    \item 
    Motivation is provided for all design choices, including algorithms, implementation choices, parameters, data sets and experimental protocols beyond metrics.
    [yes/partially/no] Partially
\end{enumerate}

\subsection*{Articles containing theoretical contributions:}
Does this paper make theoretical contributions? 
[yes/no] Yes

If yes, please complete the list below.

\begin{enumerate}
    \item All assumptions and restrictions are stated clearly and formally. 
    [yes/partially/no] Yes
    \item All novel claims are stated formally (e.g., in theorem statements). 
    [yes/partially/no] Yes
    \item Proofs of all non-trivial claims are provided in sufficient detail to permit verification by readers with a reasonable degree of expertise (e.g., that expected from a PhD candidate in the same area of AI). [yes/partially/no] Yes
    \item
    Complex formalism, such as definitions or proofs, is motivated and explained clearly.
    [yes/partially/no] Yes
    \item 
    The use of mathematical notation and formalism serves the purpose of enhancing clarity and precision; gratuitous use of mathematical formalism (i.e., use that does not enhance clarity or precision) is avoided.
    [yes/partially/no] Yes
    \item 
    Appropriate citations are given for all non-trivial theoretical tools and techniques. 
    [yes/partially/no] Yes
\end{enumerate}

\subsection*{Articles reporting on computational experiments:}
Does this paper include computational experiments? [yes/no] Yes

If yes, please complete the list below.
\begin{enumerate}
    \item 
    All source code required for conducting experiments is included in an online appendix 
    or will be made publicly available upon publication of the paper.
    The online appendix follows best practices for source code readability and documentation as well as for long-term accessibility.
    [yes/partially/no] Yes
    \item The source code comes with a license that
    allows free usage for reproducibility purposes.
    [yes/partially/no] Yes
    \item The source code comes with a license that
    allows free usage for research purposes in general.
    [yes/partially/no] Yes
    \item 
    Raw, unaggregated data from all experiments is included in an online appendix 
    or will be made publicly available upon publication of the paper.
    The online appendix follows best practices for long-term accessibility.
    [yes/partially/no] Yes
    \item The unaggregated data comes with a license that
    allows free usage for reproducibility purposes.
    [yes/partially/no] Yes
    \item The unaggregated data comes with a license that
    allows free usage for research purposes in general.
    [yes/partially/no] Yes
    \item If an algorithm depends on randomness, then the method used for generating random numbers and for setting seeds is described in a way sufficient to allow replication of results. 
    [yes/partially/no/NA] Yes
    \item The execution environment for experiments, the computing infrastructure (hardware and software) used for running them, is described, including GPU/CPU makes and models; amount of memory (cache and RAM); make and version of operating system; names and versions of relevant software libraries and frameworks. 
    [yes/partially/no] Yes
    \item 
    The evaluation metrics used in experiments are clearly explained and their choice is explicitly motivated. 
    [yes/partially/no] Yes
    \item 
    The number of algorithm runs used to compute each result is reported. 
    [yes/no] Yes
    \item 
    Reported results have not been ``cherry-picked'' by silently ignoring unsuccessful or unsatisfactory experiments. 
    [yes/partially/no] Yes
    \item 
    Analysis of results goes beyond single-dimensional summaries of performance (e.g., average, median) to include measures of variation, confidence, or other distributional information. 
    [yes/no] Yes
    \item 
    All (hyper-) parameter settings for 
    the algorithms/methods used in experiments have been reported, along with the rationale or method for determining them. 
    [yes/partially/no/NA] Yes
    \item 
    The number and range of (hyper-) parameter settings explored prior to conducting final experiments have been indicated, along with the effort spent on (hyper-) parameter optimisation. 
    [yes/partially/no/NA] Yes
    \item 
    Appropriately chosen statistical hypothesis tests are used to establish statistical significance
    in the presence of noise effects.
    [yes/partially/no/NA] Yes
\end{enumerate}

\subsection*{Articles using data sets:}
Does this work rely on one or more data sets (possibly obtained from a benchmark generator or similar software artifact)? 
[yes/no] Yes

If yes, please complete the list below.
\begin{enumerate}
    \item 
    All newly introduced data sets 
    are included in an online appendix 
    or will be made publicly available upon publication of the paper.
    The online appendix follows best practices for long-term accessibility with a license
    that allows free usage for research purposes.
    [yes/partially/no/NA] NA
    \item The newly introduced data set comes with a license that
    allows free usage for reproducibility purposes.
    [yes/partially/no] NA
    \item The newly introduced data set comes with a license that
    allows free usage for research purposes in general.
    [yes/partially/no] NA
    \item All data sets drawn from the literature or other public sources (potentially including authors' own previously published work) are accompanied by appropriate citations.
    [yes/no/NA] Yes
    \item All data sets drawn from the existing literature (potentially including authors’ own previously published work) are publicly available. [yes/partially/no/NA] Yes
    %\item All data sets that are not publicly available are described in detail.
    %[yes/partially/no/NA]
    \item All new data sets and data sets that are not publicly available are described in detail, including relevant statistics, the data collection process and annotation process if relevant.
    [yes/partially/no/NA] NA
    \item 
    All methods used for preprocessing, augmenting, batching or splitting data sets (e.g., in the context of hold-out or cross-validation)
    are described in detail. [yes/partially/no/NA] Yes
\end{enumerate}

\subsection*{Explanations on any of the answers above (optional):}
Regarding questions (2) and (3) of the "articles using data sets": we did not introduce any new dataset

\end{document}